\documentclass[12pt,letterpaper]{article}
\usepackage[top=0.75in, bottom=0.75in, left=0.75in, right=0.75in]{geometry}

\usepackage{amsmath,amsthm,amssymb,amsfonts}
\usepackage{graphicx}
\usepackage{float}
\usepackage{booktabs}
\usepackage{multirow}
\usepackage{lscape}
\usepackage{epstopdf}

\usepackage{enumitem}
\usepackage{algorithm}
\usepackage{algorithmic}

\usepackage[table]{xcolor}
\usepackage[small]{caption}
\usepackage{subcaption}
\usepackage{threeparttable}

\usepackage{microtype}
\usepackage{natbib}

\renewcommand{\textcolor}[2]{#2}

\renewcommand{\cellcolor}[2][]{ }

\renewcommand{\rowcolor}[2][]{ }

\usepackage[colorlinks]{hyperref}
\definecolor{blue}{rgb}{0.0,0.5,0.68}
\definecolor{darkblue}{rgb}{0.0,0.2,0.6}
\hypersetup{colorlinks,breaklinks,linkcolor=darkblue,urlcolor=darkblue,anchorcolor=darkblue,citecolor=blue}
\usepackage[framemethod=TikZ]{mdframed}
\usepackage[most]{tcolorbox}
\tcbset{
  promptbox/.style={
    colframe=green!70!black,
    colback=green!5,
    coltitle=black,
    fonttitle=\bfseries,
    boxrule=0.8pt,
    arc=2pt,
    left=6pt,right=6pt,top=6pt,bottom=6pt
  }
}
\newcounter{promptcounter}
\newtcolorbox{promptframed}[2][]{promptbox,title={#2},#1}

\title{TrajGPT-R: Generating Urban Mobility Trajectory with Reinforcement Learning-Enhanced Generative Pre-trained Transformer}\author{\parbox{\textwidth}{\centering%
Jiawei Wang$^{1}$,
Chuang Yang$^{1}$,
Jiawei Yong$^{2}$,
Xiaohang Xu$^{1}$,
Hongjun Wang$^{1}$,
Noboru Koshizuka$^{1}$,
Shintaro Fukushima$^{2}$,
Ryosuke Shibasaki$^{1}$,
Renhe Jiang$^{1}$\thanks{Corresponding author: \texttt{jiangrh@csis.u-tokyo.ac.jp}}%
\\
$^{1}$The University of Tokyo, Tokyo 113-0034, Japan\\
$^{2}$Toyota Motor Corporation, Tokyo 112-0004, Japan%
}}%

\date{}

\begin{document}
\maketitle

\begin{abstract}
Mobility trajectories are essential for understanding urban dynamics and enhancing urban planning, yet access to such data is frequently hindered by privacy concerns. This research introduces a transformative framework for generating large-scale urban mobility trajectories, employing a novel application of a transformer-based model pre-trained and fine-tuned through a two-phase process. Initially, trajectory generation is conceptualized as an offline reinforcement learning (RL) problem, with a significant reduction in vocabulary space achieved during tokenization. The integration of Inverse Reinforcement Learning (IRL) allows for the capture of trajectory-wise reward signals, leveraging historical data to infer individual mobility preferences. Subsequently, the pre-trained model is fine-tuned using the constructed reward model, effectively addressing the challenges inherent in traditional RL-based autoregressive methods, such as long-term credit assignment and handling of sparse reward environments. Comprehensive evaluations on multiple datasets illustrate that our framework markedly surpasses existing models in terms of reliability and diversity. Our findings not only advance the field of urban mobility modeling but also provide a robust methodology for simulating urban data, with significant implications for traffic management and urban development planning.  The implementation is publicly available at \url{https://github.com/Wangjw6/TrajGPT_R}.
\end{abstract}

\noindent\textbf{Keywords:} Urban mobility; trajectory; autoregressive generation; transformer; reward modeling; reinforcement learning

\section{Introduction}
\label{sec:1} 
Urban mobility trajectories, including vehicle routes and human movements, are critical in describing crowd dynamics in urban environments. These trajectories not only provide insights into daily travel patterns \citep{gonzalez2008understanding, alessandretti2020scales} but also reflect the underlying socio-economic interactions \citep{barbosa2021uncovering} and urban planning effectiveness \citep{gaglione2022urban}. Nonetheless, public access to such data is frequently constrained by privacy concerns, underscored by stringent data protection regulations such as the General Data Protection Regulation (GDPR) of the EU.
This limitation makes leveraging mobility data in research and practical applications challenging. Therefore, developing methods to generate reliable and diverse urban mobility trajectory data while preserving privacy is crucial for advancing academic research and enabling diverse trajectory-based applications.

Generating trajectories that accurately replicate urban dynamics is a complex task. On one hand, a sufficiently large number of generated trajectories is necessary to reveal clear mobility patterns, necessitating high model efficiency. On the other hand, the intricate urban dynamics require capturing the diverse behaviors of travelers, which demands substantial model capacity.

While previous rule-based methods \citep{isaacman2012human,zandbergen2014ensuring} can handle large-scale trajectory generation at a relatively low cost, they struggle to capture the spatial-temporal characteristics of trajectory data. Consequently, these methods are often unable to achieve a diverse and reliable generation.
Recent advancements in computational power and deep learning techniques have significantly enhanced the capacity of generative models to learn from large datasets. Notable developments include Diffusion models \citep{ho2020denoising}, Generative Adversarial Networks (GANs) \citep{goodfellow2020generative}, and Variational Autoencoders (VAEs) \citep{kingma2013auto}. These advancements have contributed to a substantial growth in effective methodologies for data generation. For example, \citet{choi2021trajgail} utilized a GAN-based framework to generate vehicle trajectories, conceptualizing trajectory generation as a sequential decision-making process with a vehicle traveling along various road segments. However, their training process is unstable due to the tricky adversarial training between the discriminator and generator. 
 Additionally, \citet{chen2021trajvae} introduced a method where each spatial-temporal point in a trajectory is encoded into a unique identifier, employing a VAE-based approach to train the mapping between latent factors and raw trajectory data. However, this approach usually suffers from too strong mode-covering behavior as the Kullback-Leibler (KL) divergence term in the loss function encourages the decoder to average over the possible outputs. 
 Moreover, in the track of diffusion-based models \citep{zhu2023difftraj,wei2024diff, zhu2024controltraj}, GPS trajectories are generated from a given noise profile through numerous denoising steps. However, too many denoising steps for a single sample generation result in higher computational costs and slower generation times compared to VAEs and GANs. Moreover, efficiently generating discrete data remains an open challenge for diffusion-based models.

 To strike a balance between generative performance and learning efficiency, large sequence models based on Transformer \citep{vaswani2017attention} and its variants (e.g., BERT \citep{devlin2018bert}, Generative Pre-trained Transformer (GPT) \citep{ethayarajh2019contextual}) have been applied to various trajectory modeling tasks \citep{lin2025uvtm, KimPSLAL23, dong2025transfusor, haydari2024mobilitygpt, dong2024enhanced, geng2023physics}. Notably, \citet{KimPSLAL23} focused on capturing key events in trajectories by modeling the temporal dependencies in human decision-making. 
\citet{liang2022trajformer} addresses the high computational cost of long trajectories by introducing an auxiliary loss to accelerate training using the supervision signals provided by all output tokens. 
\citet{haydari2024mobilitygpt} proposed a gravity-based sampling method to train the Transformer for semantic sequence similarity and leverages reinforcement learning to fine-tune LLM for optimizing trajectory generation results. 
While these initiatives demonstrate the promising capabilities of transformer-based models in trajectory data modeling, improving generalization and efficiently diversifying urban mobility trajectory generation remain significant challenges.

Unlike previous Transformer-based frameworks for trajectory generation \citep{hsu2024trajgpt}, we propose framing the task as a decision-making workflow, inspired by the Decision Transformer \citep{chen2021decision}. This approach reduces the model's parameter size by leveraging the constrained action space, which requires a smaller vocabulary and enhances trajectory generation by utilizing the sequential decision-making framework. As a result, our method improves both model efficiency and reasoning interpretability.
While previous learning methods have advanced the understanding of spatial-temporal characteristics to improve generation quality \citep{qiu2024routesformer}, they often fail to account for the distinctive trajectory preferences within urban areas, which extend beyond typical spatial-temporal crowd patterns. For example, some elderly individuals may prefer a smoother, albeit longer, route. Recognizing such preferences is crucial for capturing the true diversity of urban mobility trajectory. 
    To bridge this gap, we introduce an approach based on inverse reinforcement learning (IRL) to develop a reward model that manages to learn these distinct navigation preferences. Subsequently, we refine the pre-trained generative model by implementing a reward model-based fine-tuning scheme (RMFT). This enhancement is designed to significantly improve the model's capacity to effectively replicate diverse urban mobility trajectory patterns.

Specifically, we highlight four key contributions of our proposed framework for urban mobility trajectory generation:

\begin{itemize}
    \item \textbf{Efficient generation scheme:} We model urban mobility trajectory generation as a partially observable Markov decision process (POMDP) and leverage the Transformer architecture to enable a resource-efficient pre-trained model.
    \item \textbf{Trajectory-wise reward modeling:} An inverse reinforcement learning-based approach is introduced to calibrate a reward model, effectively capturing and explaining preferred trajectories.
    \item \textbf{Reward model-based fine-tuning:} For the first time in urban mobility trajectory generation, we enhance the pre-trained model through fine-tuning guided by an explicit reward model. This approach addresses critical challenges in Transformers, including sparse information representation and long-term credit assignment.
    \item \textbf{Validation through large-scale experiments:} Our framework is rigorously validated on multiple large-scale urban mobility trajectory datasets, showcasing significant improvements in performance and interpretability.
\end{itemize}

\section{Problem}\label{sec4}

\begin{figure}[bpht]
\centering
\includegraphics[width=0.6\textwidth]{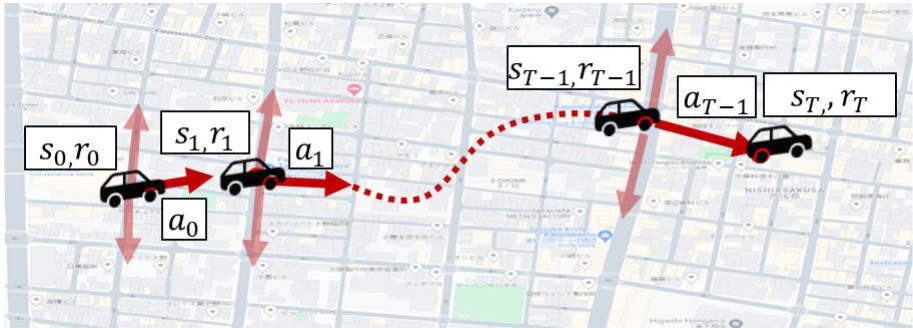}
\caption{\textbf{Trajectory generation as a sequential decision-making problem}. \small{The vehicle navigates in the urban city by making decisions to determine the downstream link at each link}}
\label{fig:problem}
\end{figure}
In this paper, we focus on large-scale urban trajectory generation tasks \textcolor{red}{(see Figure~\ref{fig:problem})}. We define a trajectory \(\tau\) as a temporally ordered sequence of road segments \( l_i \) in a road network, expressed as \(\tau = \{l_{t_0}, l_{t_1}, \ldots, l_{t_k}\}\). The generative model \(\mathcal{G}\) generates a trajectory \(\hat{\tau}\), based on predefined contextual inputs \(c\) (e.g., origin and destination), formulated as \(\hat{\tau} = \mathcal{G}(c)\).
In our proposed framework, we model the trajectory generation as a sequential decision-making problem following \citet{chen2021decision}. In this formulation, at each timestep \( t \), the agent observes the current state \( s_t \) (e.g., the current position and contextual factors like traffic conditions), selects an action \( a_t \) (e.g., determining the next downstream link segment to traverse) and receives a reward \( r_t \) representing the quality or efficiency of the chosen action (e.g., whether it arrived at the destination). The trajectory is then generated autoregressively as the agent makes decisions in each segment.

\section{Methodology}\label{sec4}
\begin{figure}[ht]
\centering
\includegraphics[width=1\textwidth]{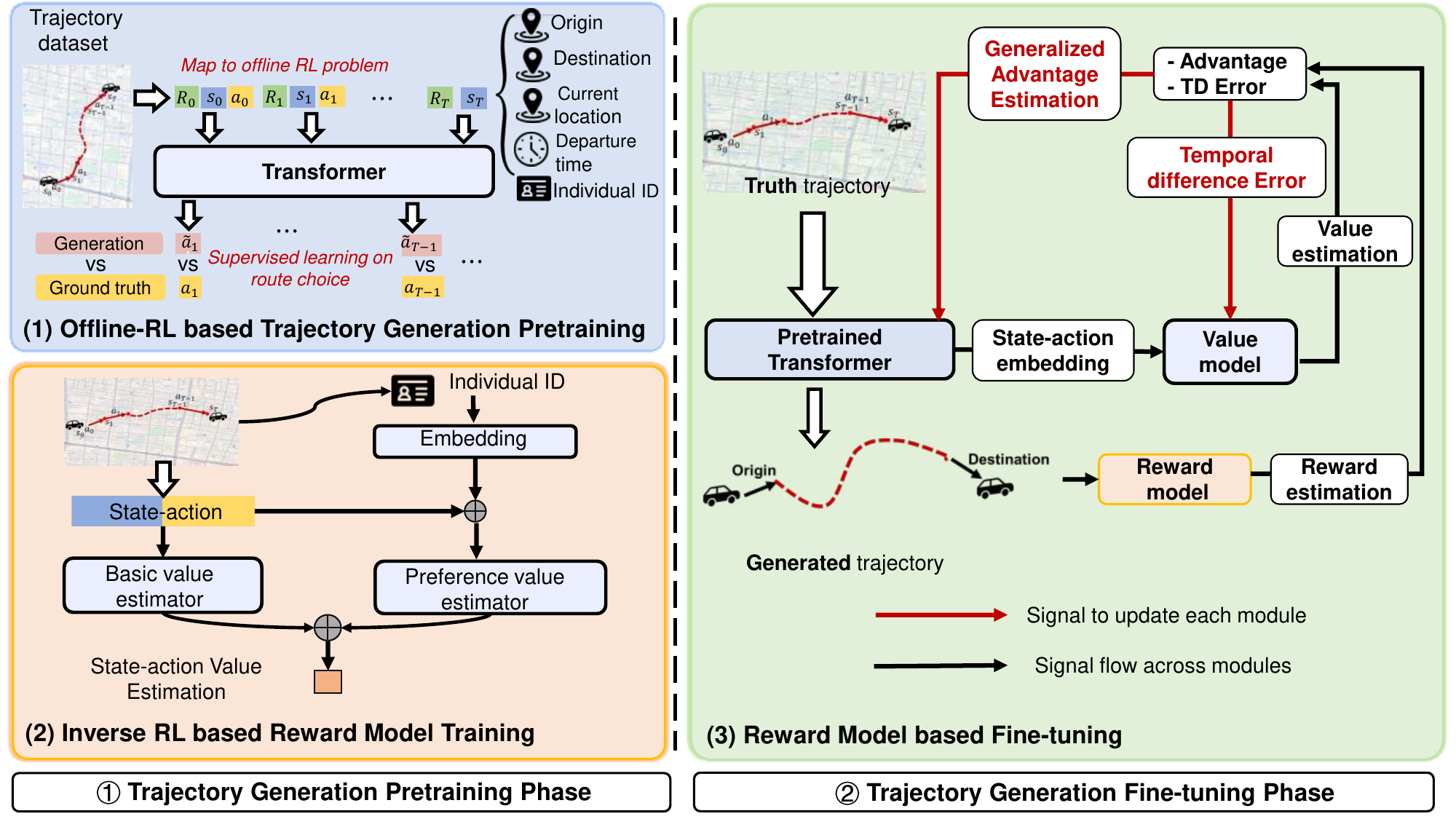}
\caption{\textbf{Our proposed Two-phase framework to enhance pretrained generative model for urban mobility trajectory generation with reinforcement learning (TrajGPT-R).} \small{\textbf{Phase 1:} A Generative \textbf{pre-trained} Transformer (GPT) is developed to acquire the general knowledge for generating urban mobility trajectory data, meanwhile a reward model is constructed using inverse reinforcement learning to capture trajectory-wise preferences. \textbf{Phase 2:} \textbf{Reward model-based fine-tuning (RMFT)} scheme is introduced to enhance the pre-trained model for better generation reliability and diversity.}}
\label{fig:overall_framework}
\end{figure}
\textcolor{red}{
Before introducing the technical details, we clarify the terminology used throughout the paper to avoid ambiguity. 
In this work, a {reward model} refers to a learned function that assigns a scalar score to a generated trajectory, which we denote as a {trajectory-wise reward} and use to capture long-term preference signals. 
During policy optimization, this trajectory-level reward is further transformed into {state-action values}, which guide token-level updates of the generative model. }
\subsection{Offline-RL based Trajectory Generation Pretraining}\label{sec4-1}
In this study, we utilize the Transformer architecture \citep{vaswani2017attention} as the backbone for trajectory modeling and follow an autoregressive generation scheme. This approach is motivated by two principal factors: Firstly, mobility trajectories share similarities with natural language, such as topological constraints in trajectories that mirror grammatical constraints in text, making the Transformer particularly effective, as demonstrated in language modeling \citep{zhao2023survey}. Secondly, the autoregressive generation scheme is suitable for sequential decision-making process modeling, where each decision is predicted on the information available at the current moment.

We employ a token-based approach to represent trajectories in our generation task based on recent advancements in sequential decision modeling with Transformer architectures \citep{chen2021decision, wang2023trajectory}. Specifically, we utilize three types of tokens\textcolor{red}{: }state tokens \(s_t\), action tokens \(a_t\), and return-to-go tokens \(R_t\) to encapsulate the decision-making context at each timestep, as shown in Figure~\ref{fig:overall_framework}-\textcolor{red}{(1)}. The autoregressive generation mechanism will ensure that each action token is generated in a manner that respects the inherent sequential dependencies of trajectories. In our study, we organize the tokens as follows:

\begin{itemize}
    \item \textbf{State token:} Each state token is designed to incorporate spatial and traffic condition information, along with individual ID for personalized context \textcolor{red}{(note that the individual ID is used to facilitate the learning of nuanced individual preferences. It is used only for reward modelling during the training stage)}. The spatial context is represented by the current link, origin link, and destination link. Furthermore, we integrate traffic-related features, such as speed and departure intervals, into the state token to inform the traffic dynamics \textcolor{red}{(note that the speed and departure interval are recorded at the first link and remain fixed during the generation process).} 
    \item \textbf{Action token:}\textcolor{red}{ Each action token corresponds to the index of the selected downstream link among the feasible outgoing links of the current link.} For example, action token 1 at link \( l \) indicates the selection of the first downstream link among all available connections from link \( l \).
    \item \textbf{Return-to-go token:} This token represents the goal of the current trajectory generation. Unlike previous studies \citep{chen2021decision},  the return-to-go in our tasks is less informative due to the uncertainty in the vehicle's route evaluation, we set the return-to-go as 1 if the vehicle is still en route to its destination and 0 once it arrives.
\end{itemize}

\begin{figure}[ht]
\centering
\includegraphics[width=1.\textwidth]{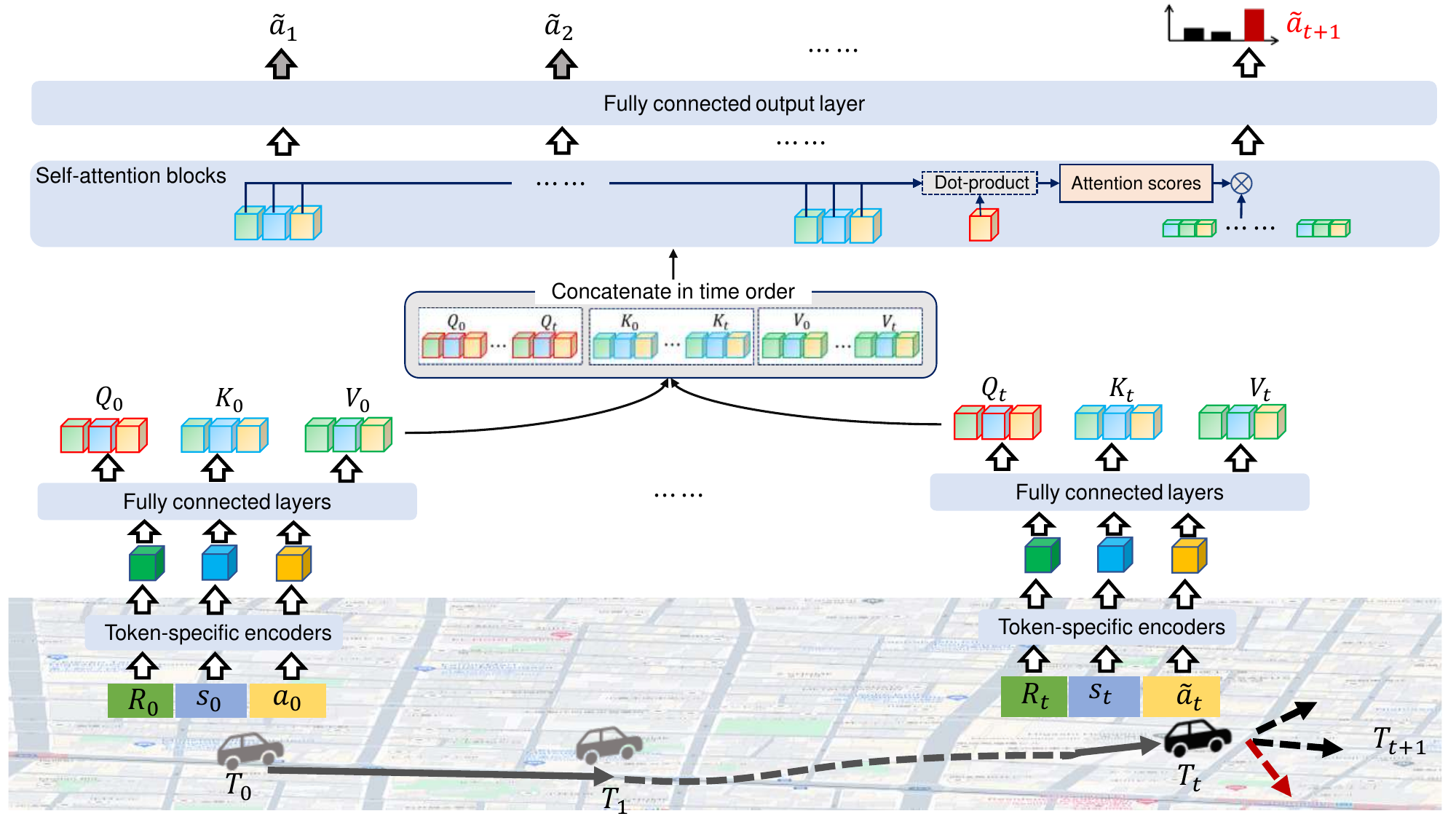}
\caption{\textbf{Autoressive decision-making process in Transformer-based mobility trajectory generation}.\small{ At each generation step \(t+1\), all preceding tokens contribute to predicting the action token at \(t+1\). The significance of each token's contribution is determined by its attention score.}}

\label{fig:generative_data_flow}
\end{figure}

To handle the proposed tokens and formulate a generative dataflow, the model architecture is as shown in Fig~\ref{fig:generative_data_flow}.\textcolor{red}{ I}n this architecture, the input tokens are first encoded into high-dimensional vectors:
\[
E_{tk} = \text{Encoder}_{tk}(x_{tk}), \quad tk \in \{s, a, R\}.
\]
The encoders for the action token ($a$) and the return-to-go token ($R$) consist of an embedding layer. Meanwhile, the encoder for the state token ($s$) comprises multiple modules. To model the encoding for the state token ($s$), each element within the state token is independently embedded to reflect its distinct physical properties, such as the link representing the spatial location and the departure time indicating temporal order.
The final state-token embedding is obtained by summing the embeddings of each element:
\[
E_s = \sum_{i} \text{Embedding}(s_i),
\]
where \( s_i \) represents the individual elements of the state token.
\[
E_{s} = \sum_{i} \text{Embedding}_{i}(x_{s_i}),
\]
where $x_{s_i}$ denotes the $i$-th element of the state token.

In succession, the concatenated embeddings from the three types of tokens at different time steps are obtained:
\[
E_{\text{cat}} = E_s \mathbin{||} E_a \mathbin{||} E_R.
\]
The query \( Q \), key \( K \), and value \( V \) matrices are then computed as:
\[
Q = E_{\text{cat}} W_Q, \quad K = E_{\text{cat}} W_K, \quad V = E_{\text{cat}} W_V.
\]
Subsequently, the attention scores are calculated by:
\[
A = \text{softmax}\left(\frac{QK^T}{\sqrt{d_k}}\right),
\]
where \( d_k \) is the dimensionality of the key vectors. Finally, the output of the attention mechanism is computed as:
\[
Z = AV.
\]
The output layer then maps this result to the final predictions through a linear transformation, optionally followed by a softmax operation for classification tasks:
\[
\hat{y} = \text{softmax}(ZW_O + b_O),
\]
where \( W_O \) and \( b_O \) are the weight matrix and bias parameters of the output layer, respectively. The softmax function is applied to produce the final probability distribution over potential actions.

As the trajectory generation in this study follows the autoregressive decision-making process, we incorporate a traditional offline RL scheme \citep{levine2020offline} in the training phase. Specifically, we are training a policy to predict action tokens based on previous tokens.
Mathematically, the objective of pretraining is formulated as a cross-entropy loss between the ground truth and the generated decision, defined as:
\begin{equation}
\mathcal{L}  = \sum_{t=1}^{T} l\left(a_t, \hat{a}_t \right),
\end{equation}
where \( \hat{a}_t = \pi_{\theta}(s_{1:t}, a_{1:t-1}, R_{1:t}) \), with \( \pi_{\theta} \) denoting the overall model parameterized by \( \theta \). Here, \( \hat{a}_t \) represents the action generated at step \( t \), given all the previously available tokens, and \( T \) denotes the total number of timesteps. The cross-entropy loss \( l \) is minimized between the generated action token \( \hat{a}_t \) and the ground truth action \( a_t \) at each timestep. For the first step, the action token is left blank since there are no prior decisions to consider.

\subsection{Inverse RL based Reward Model Construction}\label{sec4-2}

In previous studies on modeling sequential decision-making using Transformers \citep{chen2021decision,wang2023trajectory}, the return-to-go token has been introduced as a critical element for highlighting trajectory preferences. However, in the context of trajectory generation, the reward signal is often sparse and poorly defined, which results in a less informative return-to-go token. Furthermore, concerns have been raised regarding the limited capacity for long-term credit assignment in transformer architectures in RL tasks \citep{ni2024transformers}. Inspired by the benefits of the reward modeling phase in Reinforcement Learning with Human Feedback (RLHF) \citep{ouyang2022training}, this study proposes adapting this scheme to overcome the aforementioned limitations. The rationale is twofold: Firstly, the evaluation from the reward modeling effectively supplements the sparse return-to-go signal in our task. Secondly, by learning to assess the long-term effects of each action, we can leverage offline data to provide a more informative credit-assignment signal.

To this end, we propose constructing a reward model using Inverse Reinforcement Learning (IRL) from the offline data, as depicted in Figure~\ref{fig:overall_framework}-\textcolor{red}{(2)}. The proposed IRL framework is specifically designed to capture both general and individual preferences for routing evaluation, through the Basic Value Estimator (BVE) and the Preference Value Estimator (PVE), respectively. Specifically, at each decision step, we partition the available information into two categories: general information (e.g., location, origin-destination, time, and action) and individual-specific messages (e.g., the individual ID). The BVE processes the general information, while the PVE handles the individual-specific messages. The outputs of these two modules are then integrated to produce the state-action value estimates for a specific individual. These estimates are used to evaluate the benefits an individual gains from selecting a particular downstream link in the current state context.
In our study, the state-of-the-art IRL approach \citep{garg2021iq} is adapted to learn the state-action value (e.g., Q-function \( Q(s, a) \)) in our task:
\begin{align}
\mathcal{J}(Q) = & \ \mathbb{E}_{(s,a)\sim \mathcal{D}_E} \left[ 
\phi \left(Q(s,a) - \gamma \mathbb{E}_{s'\sim \mathcal{P}(\cdot|s,a)} V^*(s') \right) 
\right]   - (1-\gamma) \mathbb{E}_{\rho_0} \left[ V^*(s_0) \right],
\end{align}
where \(V^*(s)=\log \sum_{a}\exp\left(Q(s, a)\right)\), \(\mathcal{D}_E\) represents the expert demonstration (i.e., historical vehicle trajectories), and \(\phi\) is a concave function that serves as a regularizer.

\subsection{Reward Model-based Fine-tuning}\label{sec4-3}
Given the parametric reward model \( r_{\phi}(s_t, a_t) \), we obtain a trajectory-wise reward signal to evaluate the outputs from the pre-trained model. In alignment with the principles of RLHF \citep{ouyang2022training}, we propose enhancing the pre-trained model by fine-tuning it based on the previously constructed reward model \( r_{\phi}(s_t, a_t) \).

As depicted in Figure~\ref{fig:overall_framework}-\textcolor{red}{(3)}, the pre-trained model functions as the parameterized policy \(\pi_{\theta}\), predicting actions based on prior tokens, where an action \(a_t\) is sampled according to \(a_t \sim \pi_{\theta}(\cdot | \mathbf{s}_t, \mathbf{R}_t, \mathbf{a}_{t-1})\). Here, \(\mathbf{s}_t\), \(\mathbf{R}_t\), and \(\mathbf{a}_{t-1}\) represent the state, return-to-go, and previous action tokens, respectively, occurring before timestep \(t\). By learning a value model \(V_{\phi}(s_t, a_t)\) based on the reward model, we can evaluate and update the policy \(\pi_{\theta}\) using policy gradient objectives such as Generalized Advantage Estimation (GAE) \citep{schulman2015high}, defined as follows:
\begin{equation}
    A_t^{\text{GAE}(\gamma, \lambda)} = \sum_{k=0}^{\infty} (\gamma \lambda)^k \delta_{t+k},
\end{equation}
where \( V(s_t) \) is the value function estimate at state \( s_t \), \( \gamma \) is the discount factor, and \( \lambda \) is the GAE parameter that balances bias and variance in the advantage estimates. During fine-tuning, the value model is updated with Temporal Difference (TD) errors \( \delta_{t+k} \), given by: 
\[\delta_{t} = r_{\phi}(s_t, a_t) + \gamma V(s_{t+1}) - V(s_{t}).\]

Specifically, as \(A_t^{\text{GAE}(\gamma, \lambda)} \) represents a more stable advantage based on both the long-term value and the reward signal \(r_{\phi}(s_t, a_t)\) over the trajectory, we use it to formulate the final fine-tuning objective:
\begin{equation}
\max_{\pi_\theta} \mathbb{E}_{\mathbf{s}, \mathbf{R} \sim D_{\text{pref}}, a \sim \pi_{\theta}(\cdot | \mathbf{s}, \mathbf{R})} \left[ A_t^{\text{GAE}(\gamma, \lambda)} \right] - \beta \mathbb{D}_{\text{KL}}\left[\pi_{\theta}(\cdot | \mathbf{s}, \mathbf{R}) \parallel \pi_{\text{ref}}(\cdot | \mathbf{s}, \mathbf{R}) \right].
\end{equation}
Note that we use the penalized term \( \beta \mathbb{D}_{\text{KL}}\left[\pi_{\theta}(\cdot | \mathbf{s}, \mathbf{R}) \parallel \pi_{\text{ref}}(\cdot | \mathbf{s}, \mathbf{R}) \right] \) to prevent the updated policy from evolving too abruptly, where \(\beta \) is the hyper-parameter weight.

\textcolor{red}{
Instead of directly optimizing a KL-regularized reinforcement learning objective, we empirically observe that a supervised fine-tuning formulation, in which the reward signal is incorporated as an auxiliary weighting term, yields more stable optimization. Specifically, we optimize the following objective:
\begin{equation}
\max_{\pi_\theta} 
\alpha \,
\mathbb{E}_{\mathbf{s}, \mathbf{R} \sim D_{\text{pref}},\, a \sim \pi_{\theta}(\cdot \mid \mathbf{s}, \mathbf{R})}
\left[ A_t^{\text{GAE}(\gamma, \lambda)} \right]
- \mathcal{L},
\end{equation}
where $\mathcal{L}$ denotes the supervised loss of the generative Transformer, and $\alpha$ controls the strength of the reward-guided update. In our experiments, we set $\alpha = 0.02$ to enforce conservative updates and ensure stable fine-tuning.}

This fine-tuning scheme is expected to improve performance in two primary ways: 
First, the reward model, trained using IRL, adaptively extracts preference-based information to provide immediate reward signals \( r_{\phi}(s_t, a_t) \) for each generation. 
Second, by incorporating reward signals with online updating of the value model, the derived fine-tuning objective accounts for the long-term effects at each generation step, addressing the long-term credit assignment gap typically faced by transformer-based pre-trained models in decision-making tasks \citep{ni2024transformers}.

\section{Results}\label{sec2}

\subsection{Tasks and Datasets}
In this paper, we focus on large-scale trajectory generation tasks in the context of urban areas. We define a trajectory \(\tau\) as a temporally ordered sequence of road segments \( l_i \) in a road network, expressed as \(\tau = \{l_{t_0}, l_{t_1}, \ldots, l_{t_k}\}\). The generative model \(\mathcal{G}\) generates a trajectory \(\hat{\tau}\), based on predefined contextual inputs \(c\) (such as origin and destination points), formulated as \(\hat{\tau} = \mathcal{G}(c)\).

In our proposed framework, we model the trajectory generation as a sequential decision-making problem following \citet{chen2021decision}. In this formulation, at each timestep \( t \), the agent observes the current state \( s_t \) (which could include information such as the current position and contextual factors like traffic conditions), selects an action \( a_t \) (e.g., determining the next link segment downstream to traverse), and receives a reward \( r_t \) representing the quality or efficiency of the chosen action (e.g., whether arrived at the destination). The trajectory is then generated autoregressively, as the agent progresses through each segment, making decisions at every step.


We consider multiple large-scale urban mobility trajectory datasets to validate the generation performance and applicability, including:

\noindent \textbf{Toyota Dataset}. This dataset consists of 295,488 GPS trajectories collected from Toyota vehicles operating in the Tokyo metropolitan area, covering the period from October 1 to December 31, 2021. The study area includes 290,016 road links. For the trajectory generation task, we define the action space as consisting of 9 possible downstream links, representing the maximum number of candidate links available at each decision step during trajectory generation. The final dataset data processing comprises 295,488 trajectories.

\noindent \textbf{T-Drive Dataset}. \citep{yuan2010t} This dataset contains GPS trajectories from 10,357 taxis in Beijing, recorded between February 2 and February 8, 2008. In our analysis, we divide the area into 1 km grids and resample decision steps within each trajectory every 10 minutes. This results in a research area comprising 13,433 grids. Grid shifts were modeled as actions, resulting in an action space of 9 possible moves (e.g., action 1 represents a 1-grid shift in horizontal and vertical directions). The final dataset after data processing comprises 367,852 trajectories.

\noindent \textbf{Porto Taxi Dataset}. \citep{moreira2013predicting} This dataset includes GPS trajectories of 441 taxis operating in Porto, Portugal, recorded over one year (from July 1, 2013, to June 30, 2014). For our analysis, we divided the Porto metropolitan area into 200 m$^2$ grids and resampled the decision steps within each trajectory to occur every 10 minutes. This approach results in a research area consisting of approximately 5524 grids. Grid transitions were treated as actions, leading to an action space of 9 possible moves (e.g., action 1 represents a shift by one grid in both horizontal and vertical directions). The final dataset after data processing comprises over 0.7 million trajectories.


\subsection{Evaluation Metrics}
To comprehensively evaluate the generated results, eight generation metrics have been used to evaluate the performance of the model. These include Jaccard similarity (Jac), Cosine similarity (Cos), BLEU, Jensen-divergence of link distribution (L-JSD), and Jensen-divergence of connection distribution (C-JSD). These metrics are designed to gauge the reliability of the results from a microscopic perspective (i.e. Jac, Cos, and BLEU) and an aggregated perspective (i.e., L-JSD and C-JSD). In addition, the diversity of generation within the urban context is also assessed, particularly through the unigram entropy (UE) and bigram entropy (BE), which measure the diversity of individual links and transitions between links, respectively. The details of these metrics are introduced in the Appendix~\ref{secA1}.

\subsection{Model Configurations}
All experiments were performed using Python 3.11.8. The deep learning methods are implemented using PyTorch 2.5.0. We run all experiments on a server running Ubuntu 22.04.4, equipped with four NVIDIA RTX A6000 GPUs. Additional details of model configurations are available in the Appendix~\ref{secA2}.

\subsection{Baselines}
To evaluate the efficacy of our proposed generation framework, we compared it against a diverse set of baseline models that utilize different generation schemes or architectures. The selected baselines include: a statistical method, \textbf{Markov} \citep{korolyuk1975semi}; a GAN-based method, \textbf{TrajGAIL} \citep{choi2021trajgail}; a VAE-based generation method, \textbf{TrajVAE} \citep{chen2021trajvae}; an IRL-based training method, \textbf{IQL} \citep{garg2021iq}; a diffusion model-based generation approach, \textbf{D3PM} \citep{austin2021structured}; and two ablation baselines, \textbf{TrajGPT} and \textbf{TrajGPT-DPO} \citep{rafailov2024direct}. Here, \textbf{TrajGPT} refers to the pre-trained phase (i.e., phase 1), while \textbf{TrajGPT-DPO} refers to the fine-tuning of the pre-trained model using the fine-tuning scheme without an explicit reward model. The latter two are regarded as ablation baselines for the proposed \textbf{TrajGPT-R}, in which RMFT is adopted as the fine-tuning scheme.

\subsection{Performance Evaluation}
To comprehensively evaluate the performance of our trajectory generation methods, we conducted experiments across three benchmark datasets: Toyota, T-Drive, and Porto. Since these datasets originate from different geographical areas and represent distinct demographics (i.e., the general public for the Toyota dataset and taxi drivers for T-Drive and Porto), this diversity allows us to thoroughly assess the generalization capabilities of our proposed framework.
 
The experimental results are reported from Table~\ref{tab:toyota_metric} to Table~\ref{tab:porto_metric}.
For each dataset, we generate 5,000 trajectories for evaluation. Upon evaluation across all metrics, our proposed framework (e.g., TrajGPT-R) demonstrates superior performance in generating diverse and accurate trajectories. Specifically, it achieves the highest reliability scores, including a Jaccard index of 0.524, a cosine similarity of 0.575, and a BLEU score of 0.383. It also records the lowest values in L-JSD (0.016) and C-JSD (0.042), indicating minimal distribution divergence. Additionally, the framework attains high entropy scores, with a UE of 14.85 and a BE of 14.82.
Similar trends can be observed in the T-Drive dataset, with TrajGPT-R slightly outperforming other models. It achieved the highest Jac score of 0.635 and competitive results in Cos (0.570) and BLEU (0.345). Both TrajGPT-R and TrajGPT-DPO demonstrate excellent efficiency, tying for the lowest L-JSD and C-JSD scores, which validates the necessity of fine-tuning.
The performance across the Porto dataset is consistent with the other datasets, with TrajGPT-R consistently outperforming other models in most metrics. This is particularly notable given the complex urban dynamics represented in the Porto data. 

The experimental results validate the effectiveness of the proposed TrajGPT-R framework, particularly its ability to handle diverse urban contexts and its superiority in balancing accuracy with diversity in trajectory generation. The enhancements achieved during the fine-tuning phase, supported by the reward model, significantly contribute to its performance, surpassing both traditional and advanced trajectory modeling techniques. Notably, as the three datasets exhibit different trip dynamics (e.g., general public preferences in the Toyota dataset versus taxi driver behaviors in the T-Drive and Porto datasets), the TrajGPT-R framework demonstrates that through fine-tuning with an explicit reward model, we can effectively handle urban mobility trajectory generation across various scenarios.

\begin{table}[ht]
\centering
\caption{{Method Comparison across Metrics for the Toyota Dataset}}
\setlength{\tabcolsep}{2pt} 
\renewcommand{\arraystretch}{1.1} 
\begin{scriptsize} 
\begin{tabular}{l|ccccc|cc}
\toprule
\textbf{Method} & \multicolumn{5}{c|}{\cellcolor{blue!15}\textbf{Reliability}} & \multicolumn{2}{c}{\cellcolor{red!15}\textbf{Diversity}} \\
\cmidrule(lr){2-6} \cmidrule(lr){7-8}
& \textbf{Jac(↑)} & \textbf{Cos(↑)} & \textbf{BLEU(↑)} & \textbf{L-JSD(↓)} & \textbf{C-JSD(↓)} & \textbf{UE(↑)} & \textbf{BE(↑)} \\
\midrule
\textbf{Markov} & 0.198 & 0.291 & 0.008 & 0.340 & 0.692 & 13.40 & 13.75 \\
\textbf{TrajVAE} & 0.181 & 0.271 & 0.018 & 0.056 & 0.174 & 12.31 & 12.49 \\
\textbf{TrajGAIL} & 0.124 & 0.206 & 0.001 & 0.072 & 0.234 & 12.22 & 12.61 \\
\textbf{D3PM} & 0.173 & 0.225 & 0.021 & 0.692 & 0.691 & 12.02 & 12.24\\
\textbf{IQL} & 0.236 & 0.271 & 0.072 & 0.026 & 0.075 & 14.17 & 14.28 \\
\rowcolor{orange!30}\textbf{TrajGPT} & 0.390 & 0.455 & 0.225 & 0.028 & 0.070 & 14.07 & 14.41 \\
\rowcolor{orange!30}\textbf{TrajGPT-DPO} & 0.499 & 0.556 & 0.341 & 0.021 & 0.052 & 14.44 & 14.75 \\
\rowcolor{green!30}\textbf{TrajGPT-R} & \textbf{0.524} & \textbf{0.575} & \textbf{0.383} & \textbf{0.016} & \textbf{0.042} & \textbf{14.85} & \textbf{14.82} \\
\midrule
\end{tabular}
\end{scriptsize}
\label{tab:toyota_metric}
\end{table}

\begin{table}[ht]
\centering
\caption{Method Comparison Across Metrics for the T-Drive Dataset}
\setlength{\tabcolsep}{2pt} 
\renewcommand{\arraystretch}{1.1} 
\begin{scriptsize} 
\begin{tabular}{l|ccccc|cc}
\toprule
\textbf{Method} & \multicolumn{5}{c|}{\cellcolor{blue!15}\textbf{Reliability}} & \multicolumn{2}{c}{\cellcolor{red!15}\textbf{Diversity}} \\
\cmidrule(lr){2-6} \cmidrule(lr){7-8}
& \textbf{Jac(↑)} & \textbf{Cos(↑)} & \textbf{BLEU(↑)} & \textbf{L-JSD(↓)} & \textbf{C-JSD(↓)} & \textbf{UE(↑)} & \textbf{BE(↑)} \\
\midrule
\textbf{Markov}     & 0.225 & 0.428 & 0.000  & 0.007 & 0.137 & 3.66  & 3.63 \\
\textbf{TrajVAE}    & 0.200 & 0.405 & 0.054  & 0.054 & 0.144 & 8.27& 9.00 \\
\textbf{TrajGAIL}   & 0.172 & 0.274 & 0.000  & 0.166 & 0.328 & 8.06 & 9.06 \\
\textbf{D3PM}       & 0.231 & 0.246 & 0.127  & 0.691 & 0.691 & 7.89  & 8.65 \\
\textbf{IQL}        & 0.314 & 0.374 & 0.195  & 0.005 & 0.010 & 8.40 &9.81 \\
\rowcolor{orange!30}\textbf{TrajGPT} & 0.617 & 0.567 & 0.317  & 0.005 & 0.014 & 8.06 & 9.87  \\
\rowcolor{orange!30}\textbf{TrajGPT-DPO}        & 0.612 & \textbf{0.573} & 0.333  & 0.005 & \textbf{0.011} & 8.49  & 10.12 \\
\rowcolor{green!30}\textbf{TrajGPT-R}        & \textbf{0.635} & 0.570 & \textbf{0.345}  & \textbf{0.005} & 0.013 &  \textbf{8.57}  & \textbf{10.22} \\
\bottomrule
\end{tabular}
\end{scriptsize}
\label{tab:drive_metric}
\end{table}

\begin{table}[ht]
\centering
\caption{{Method Comparison across Metrics for the Porto Dataset}}
\setlength{\tabcolsep}{2pt} 
\renewcommand{\arraystretch}{1.1} 
\begin{scriptsize} 
\begin{tabular}{l|ccccc|cc}
\toprule
\textbf{Method} & \multicolumn{5}{c|}{\cellcolor{blue!15}\textbf{Reliability}} & \multicolumn{2}{c}{\cellcolor{red!15}\textbf{Diversity}} \\
\cmidrule(lr){2-6} \cmidrule(lr){7-8}
& \textbf{Jac(↑)} & \textbf{Cos(↑)} & \textbf{BLEU(↑)} & \textbf{L-JSD(↓)} & \textbf{C-JSD(↓)} & \textbf{UE(↑)} & \textbf{BE(↑)} \\
\midrule
\textbf{Markov}     & 0.080 & 0.332 & 0.099  & 0.077 & 0.209 & 6.71  & 6.71 \\
\textbf{TrajVAE}    & 0.080 & 0.320 & 0.158  & 0.093 & 0.226 & 6.63 & 6.71\\
\textbf{TrajGAIL}   & 0.102 & 0.247 & 0.050 & 0.021 & 0.035 & 8.46 & 10.47 \\
\textbf{D3PM}       &  0.083 &  0.333 &  0.259 &  0.111 & 0.225 & 6.81& 6.82 \\
\textbf{IQL}        & 0.214 & 0.259 & 0.215  & 0.023 & 0.042 & 7.77 & 8.52 \\
\rowcolor{orange!30}\textbf{TrajGPT} &0.319 & 0.350 & 0.322  & 0.023 & 0.042 & 8.13& 9.17  \\
\rowcolor{orange!30}\textbf{TrajGPT-DPO}        & 0.337 & 0.353 &  0.332  & 0.024 & 0.040 & 8.06 & 9.06 \\
\rowcolor{green!30}\textbf{TrajGPT-R}       & \textbf{0.522} & \textbf{0.470} & \textbf{0.432} & \textbf{0.013} & \textbf{0.032} & \textbf{10.13} & \textbf{10.75}\\
\bottomrule
\end{tabular}
\end{scriptsize}
\label{tab:porto_metric}
\end{table}

\begin{figure}[htbp]
    \centering
\begin{subfigure}[b]{0.45\textwidth}
        \centering
        \includegraphics[width=\textwidth]{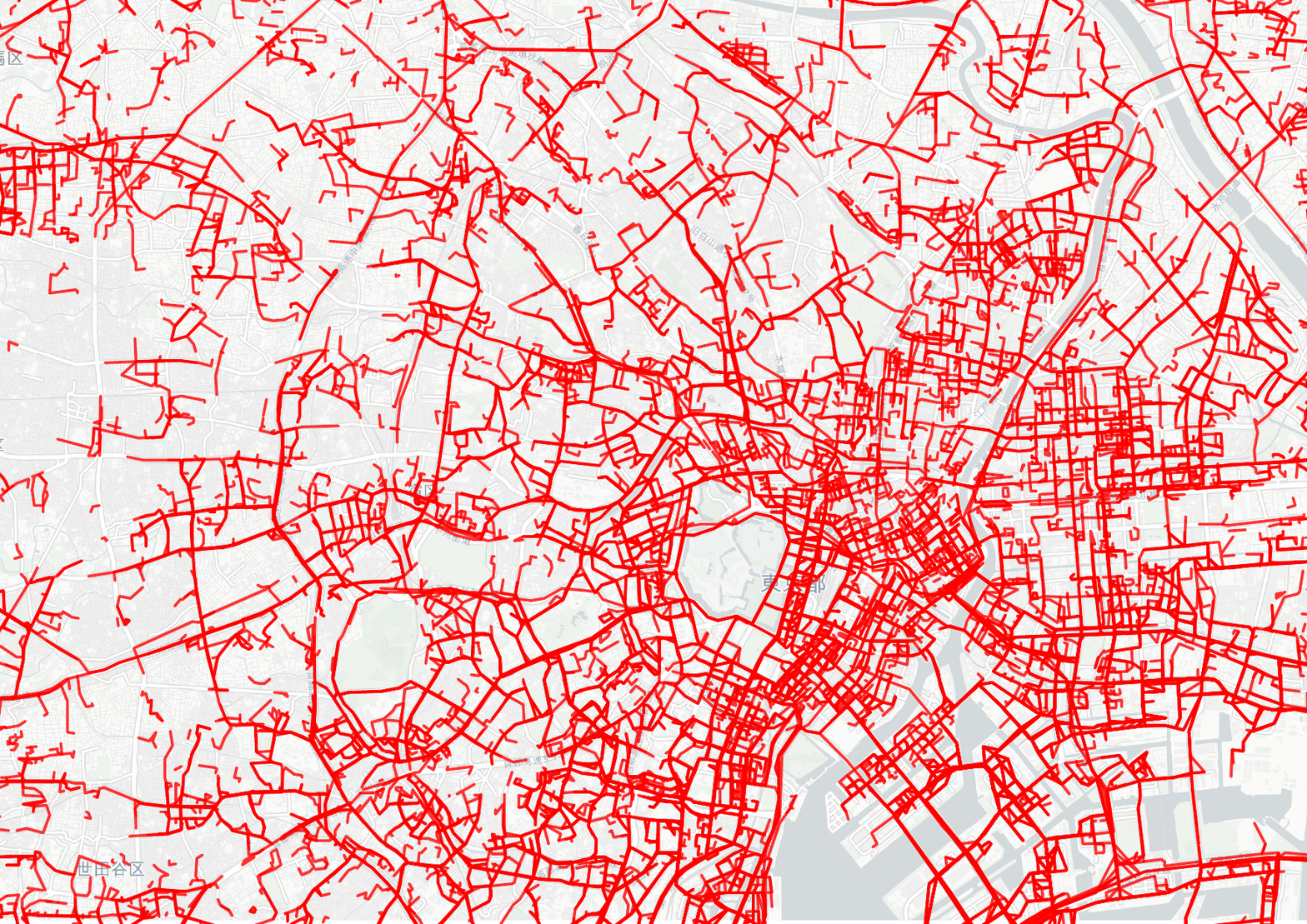}
        \caption{Ground truth}
        \label{fig:individual_ini}
    \end{subfigure}
    \hfill
    \begin{subfigure}[b]{0.45\textwidth}
        \centering
        \includegraphics[width=\textwidth]{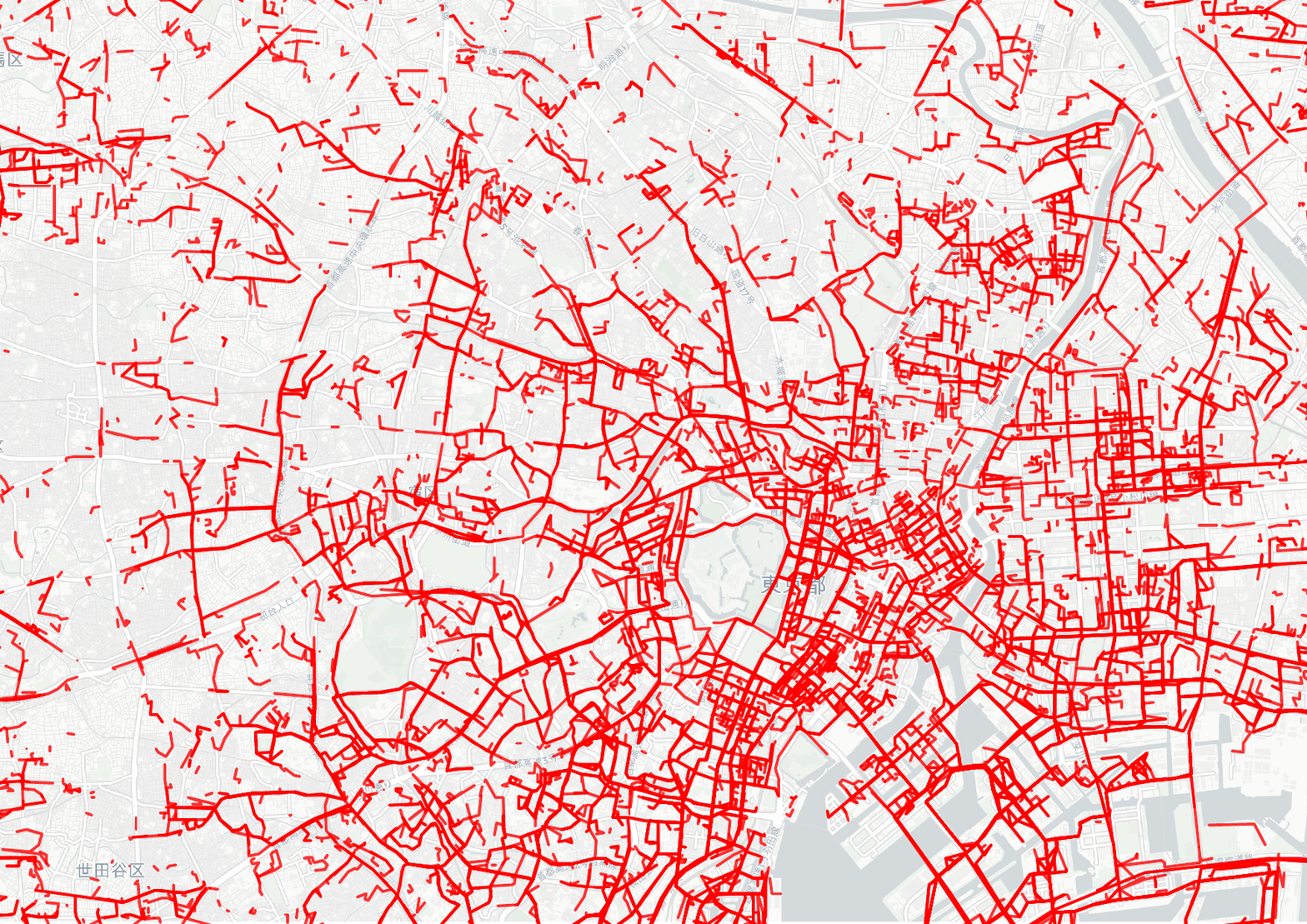}
        \caption{TrajGPT}
        \label{fig:individual_pre}
    \end{subfigure}

    \begin{subfigure}[b]{0.45\textwidth}
        \centering
        \includegraphics[width=\textwidth]{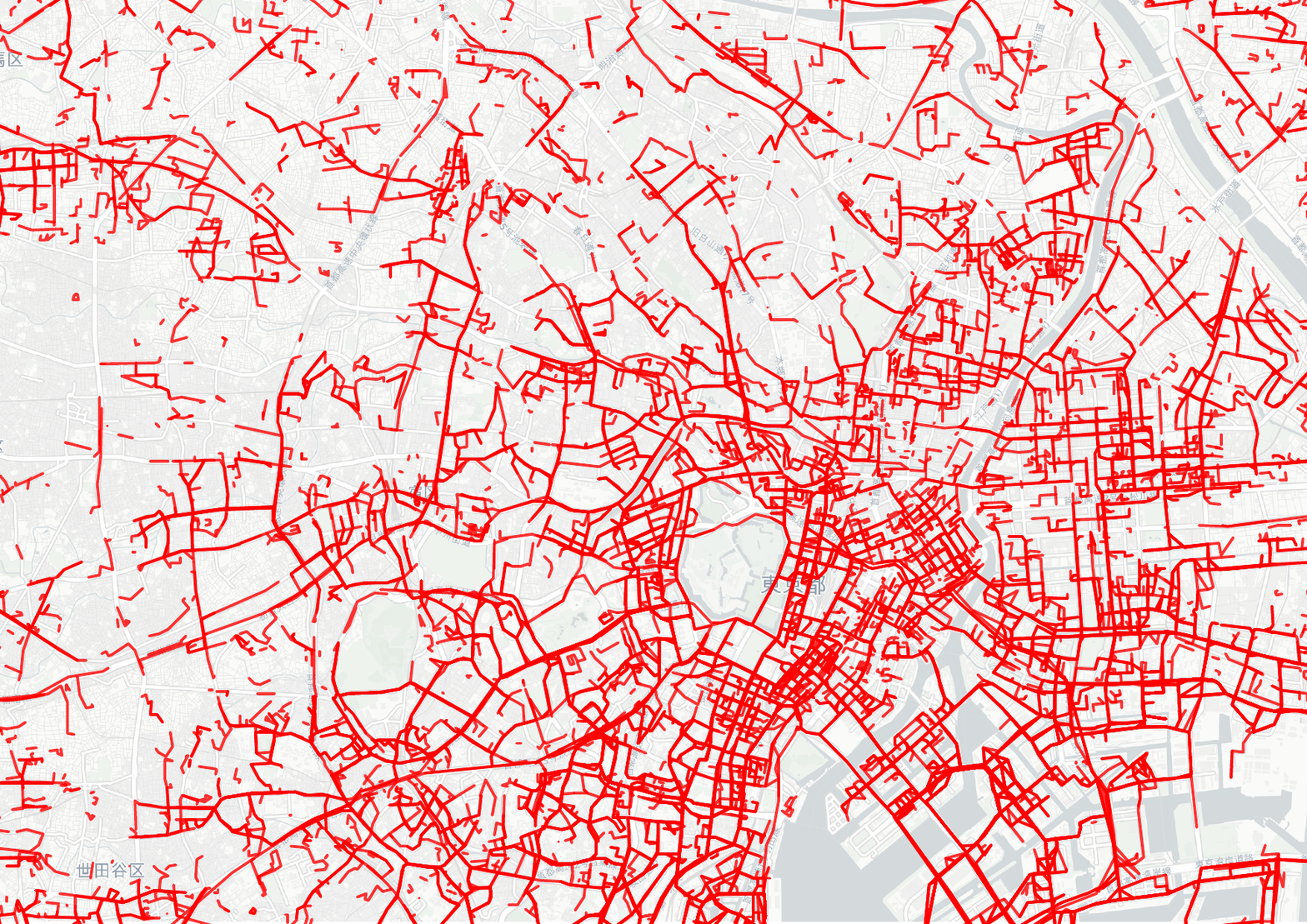}
        \caption{TrajGPT-DPO}
        \label{fig:individual_dpo}
    \end{subfigure}
    \hfill
    \begin{subfigure}[b]{0.45\textwidth}
        \centering
        \includegraphics[width=\textwidth]{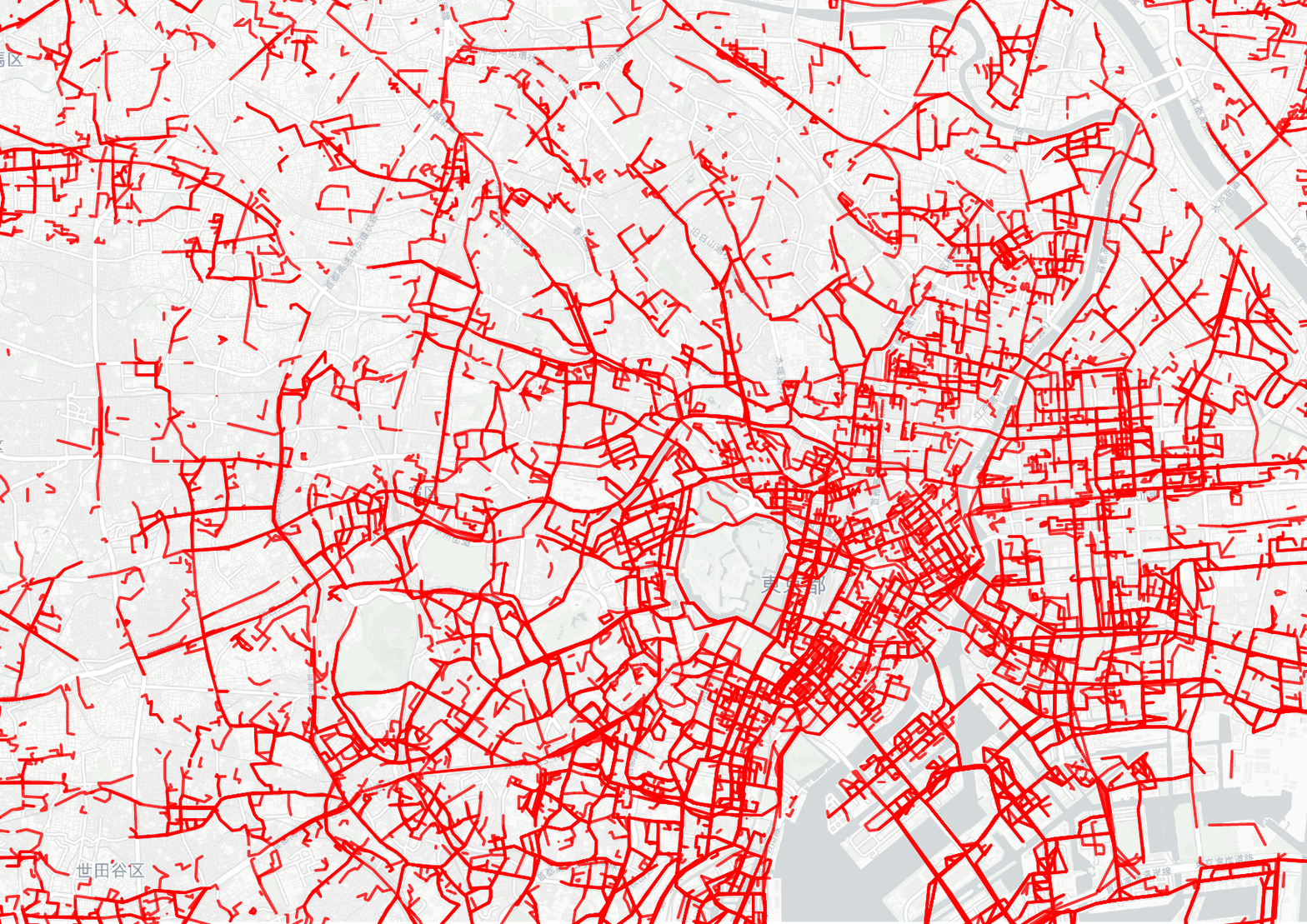}
        \caption{TrajGPT-R}
        \label{fig:individual_rlhf}
    \end{subfigure}
    \caption{\textbf{Visualization of generated and the ground-truth trajectories based on Toyota Dataset within the core area of Tokyo, Japan.} \small{The trajectories are drawn in red. We highlight different training phases. Compared with the ground truth, TrajGPT can present satisfying reproduction performance. Moreover, the fine-tuning phase (e.g., TrajGPT-DPO and TrajGPT-R) can improve the model's generalization ability.}}
    \label{fig:toyota_full_traj}
\end{figure}

For a more intuitive examination of the generation results, we visualize the trajectories generated by our framework and compare them with those from ablation baselines. Fig~\ref{fig:toyota_full_traj} displays 5,000 generated and ground-truth trajectories based on the Toyota Dataset within the core area of Tokyo, Japan. We observe that the fine-tuning phase through DPO or RMFT significantly enhances the TrajGPT capability to accurately generate trajectories in sparsely populated areas (e.g., the north area of the map). This improvement underscores the positive impact of the RL fine-tuning scheme on model generalization, aligning with the findings reported in \cite{tajwar2024preference}.

\begin{figure}[htbp]
    \centering
\begin{subfigure}[b]{0.45\textwidth}
        \centering
        \includegraphics[width=\textwidth]{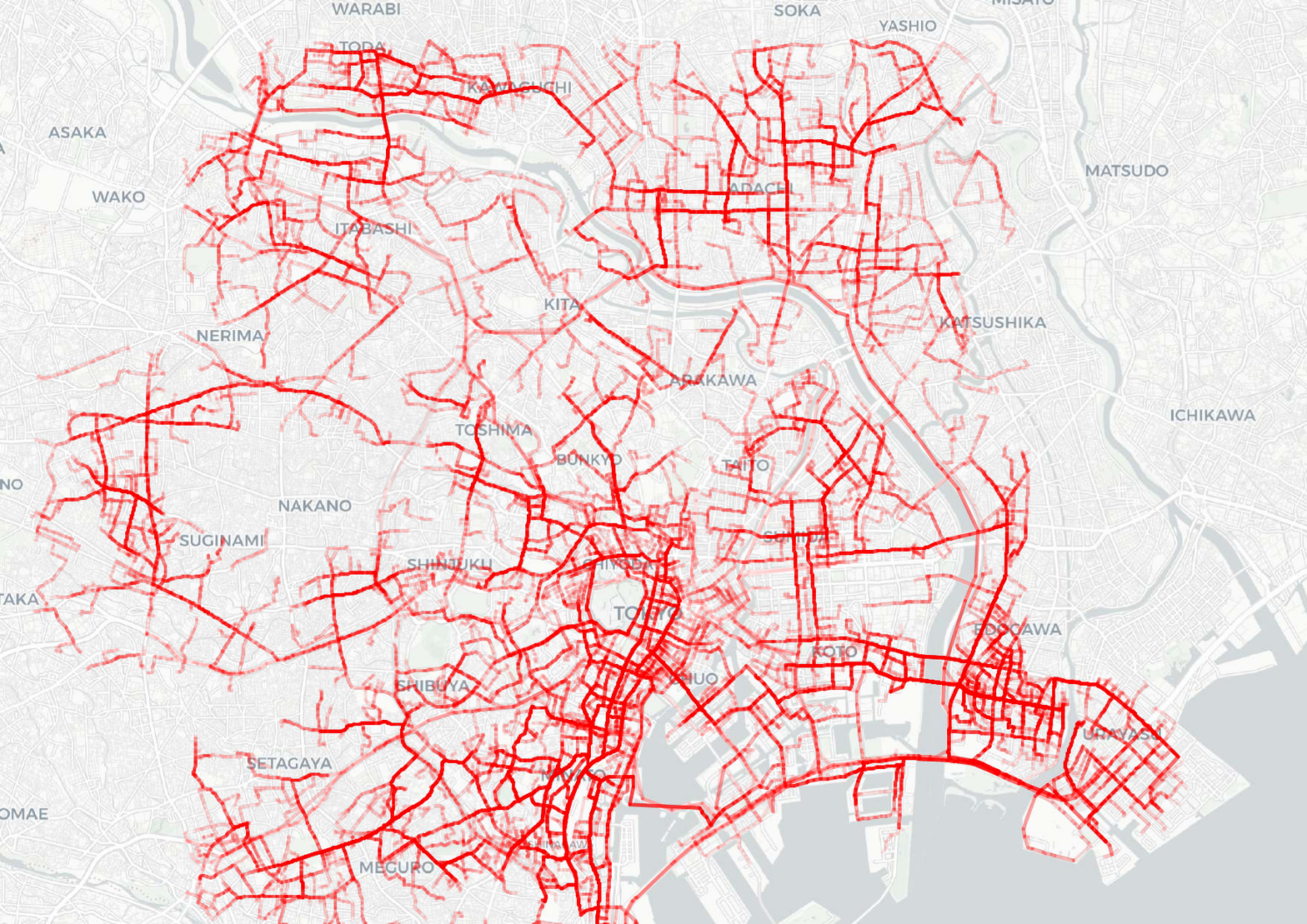}
        \caption{Initialization}
        \label{fig:individual_ini}
    \end{subfigure}
    \hfill
    \begin{subfigure}[b]{0.45\textwidth}
        \centering
        \includegraphics[width=\textwidth]{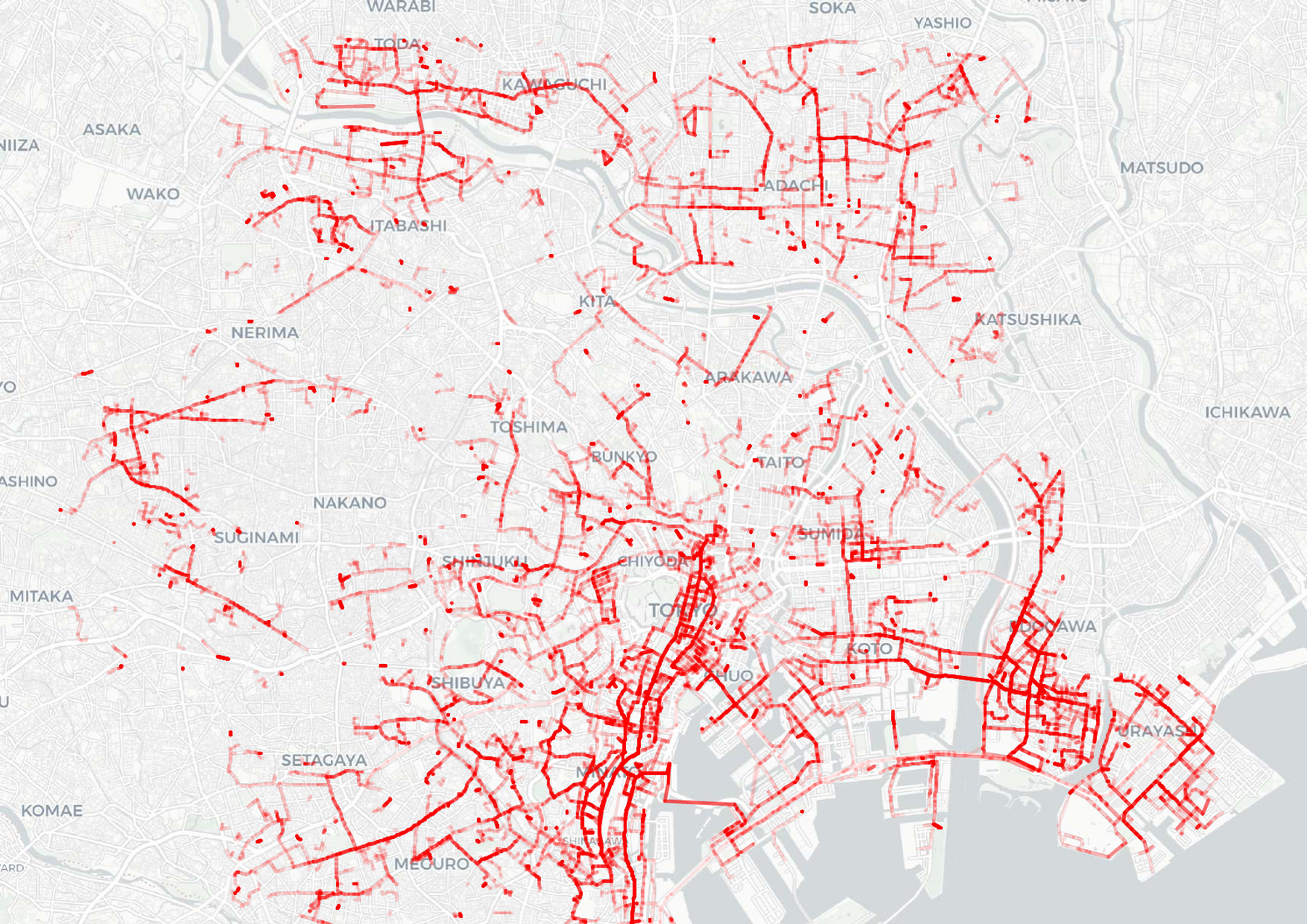}
        \caption{TrajGPT}
        \label{fig:individual_pre}
    \end{subfigure}

    \begin{subfigure}[b]{0.45\textwidth}
        \centering
        \includegraphics[width=\textwidth]{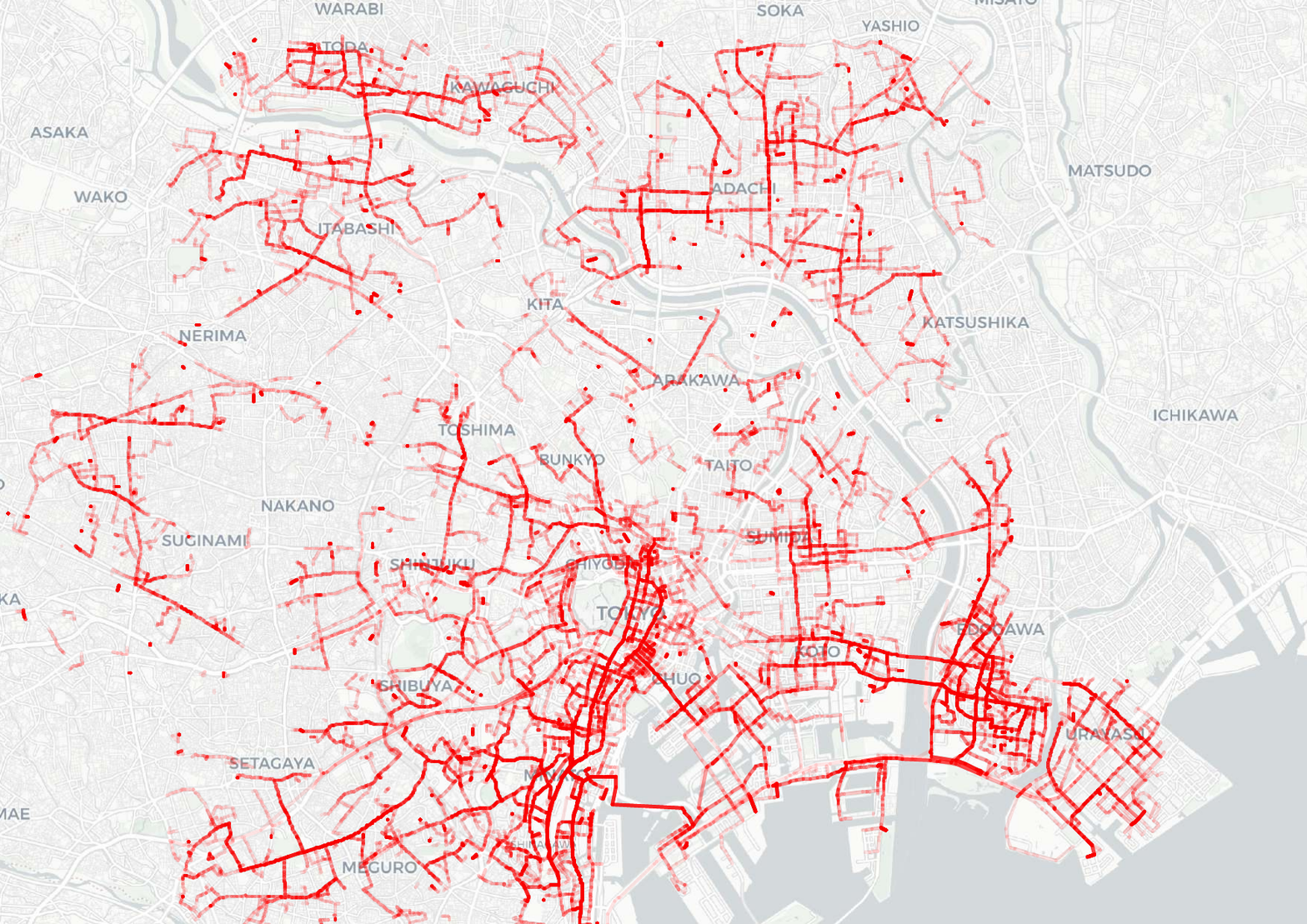}
        \caption{TrajGPT-DPO}
        \label{fig:individual_dpo}
    \end{subfigure}
    \hfill
    \begin{subfigure}[b]{0.45\textwidth}
        \centering
        \includegraphics[width=\textwidth]{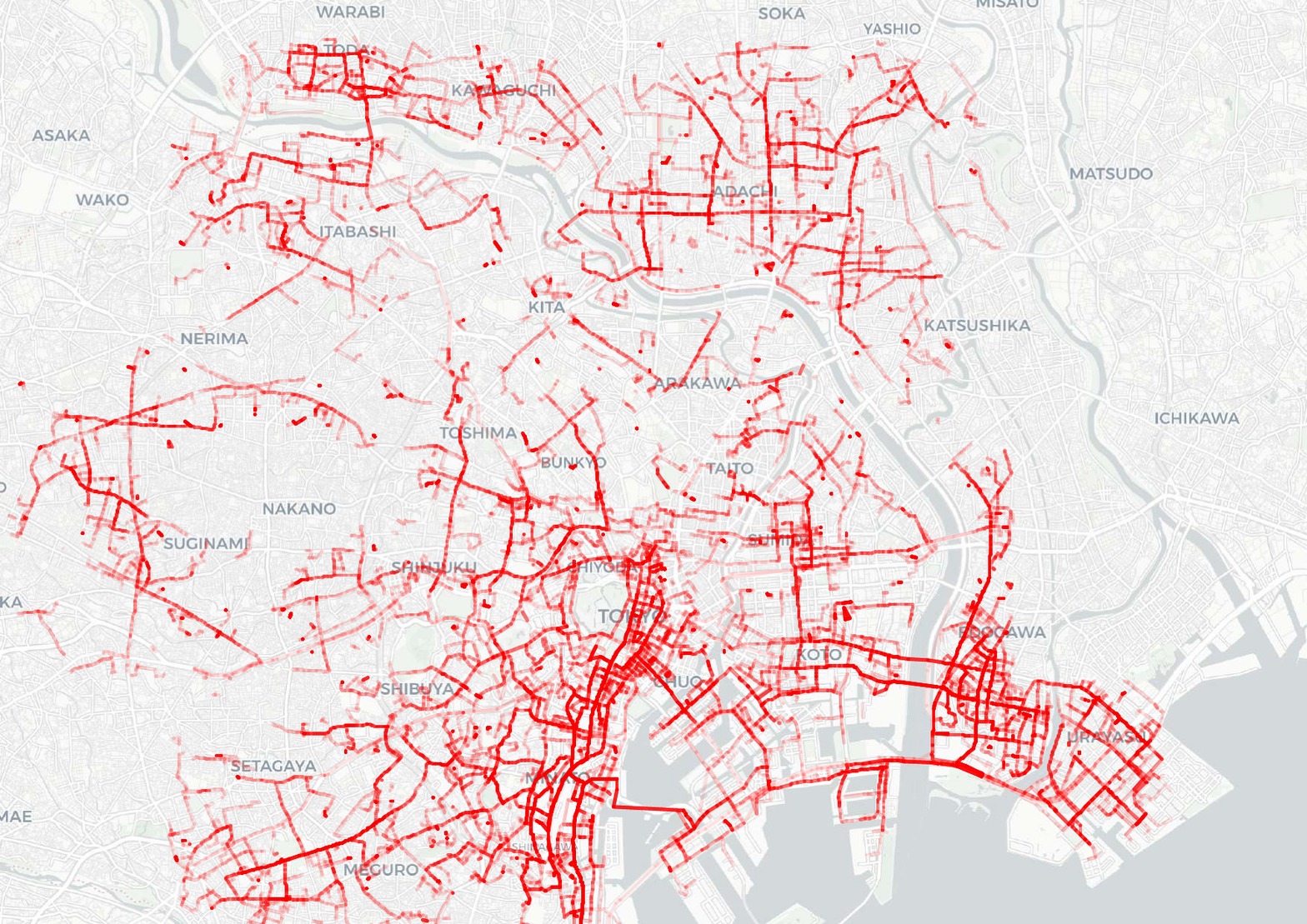}
        \caption{TrajGPT-R}
        \label{fig:individual_rlhf}
    \end{subfigure}
    \caption{\textbf{Long trajectory visualization.} \small{We specially select and visualize the trajectories consisting of more than 50 links. The pre-trained model presents a worse reproduction. RMFT especially outperforms the other two phases in long trajectory modeling (see e.g., the western part of the map). }}
    \label{fig:long_vis}
\end{figure}

Furthermore, as illustrated in Fig~\ref{fig:long_vis}, we analyze the long trajectory visualizations. It is evident that TrajGPT-R more closely replicates the ground truth trajectory compared to the other two ablation baselines. This superior performance likely stems from explicit reward modeling, which enhances the transformer's ability to assign credit accurately in decision-making tasks, thereby improving the model's efficacy in generating longer trajectories.

\begin{figure}[htbp]
\centering
\includegraphics[width=1.0\textwidth]{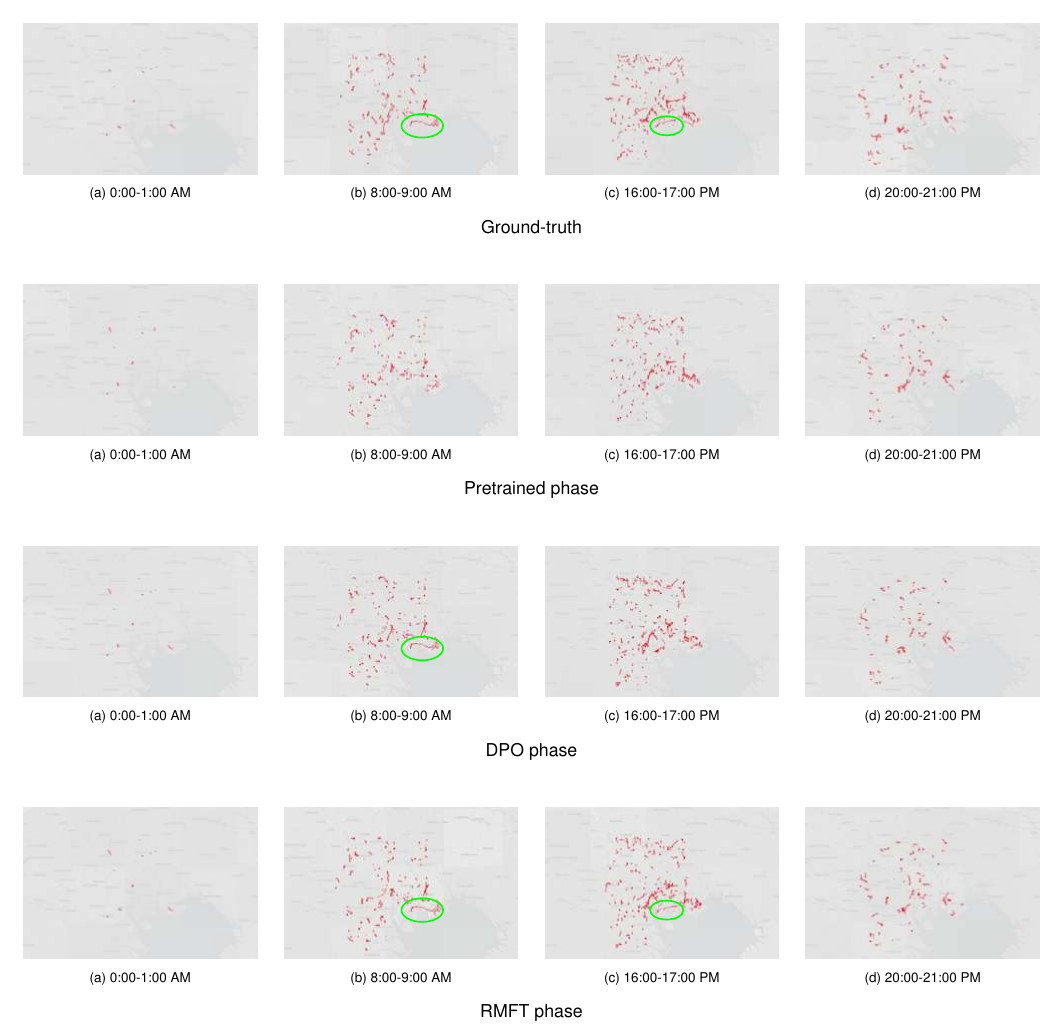}
\caption{\textbf{Temporal variation in trajectory generation.} We compare generated trajectories across four representative time periods—morning non-peak, morning peak, afternoon peak, and afternoon non-peak—to illustrate temporal dynamics in urban mobility. Fine-tuning strategies such as DPO and RMFT enhance the pretrained model's ability to capture subtle trajectory patterns observed in real-world data (i.e., highlighted in green circles). Notably, the proposed RMFT method more effectively captures nuanced mobility behaviors, resulting in more temporally consistent and realistic trajectory generation.}
\label{fig:temp_vis}
\end{figure}

The robustness of the generation can be further examined along the temporal dimension, as shown in Fig~\ref{fig:temp_vis}. This analysis reveals how effectively the model captures variations over time and adapts to varying traffic conditions. Specifically, TrajGPT-R demonstrates a consistent ability to generate trajectories that align closely with real-world temporal patterns. These results underscore the model's capacity for handling real-time variabilities and maintaining reliability across extended periods, which is promising for practical data generation for urban planning or traffic management.

\subsection{Interpretability Analysis}
In this section, we discuss the mechanism of the proposed framework for mastering trajectory generation through an intuitive and interpretative approach. Specifically, by integrating explicit individual modeling via individual ID embeddings—which potentially encode personal preferences—and employing an autoregressive decision-making scheme, we address the following key questions:

\begin{itemize}
    \item \textbf{Q1:} How do individual embeddings evolve in different training stages?
    \item \textbf{Q2:} What is the contribution of each input token type to trajectory generation?
    \item \textbf{Q3:} How does the autoregressive decision-making scheme utilize each token?
\end{itemize}

Without loss of generality, we utilize the Toyota dataset for our analysis and demonstrations. More than 50,000 trajectories are generated through a sequential four-phase process using our foundational transformer model. These phases include Initialization, the pre-trained model (TrajGPT), and two distinct fine-tuned models, TrajGPT-DPO and TrajGPT-R, employing different tuning schemes: DPO and RMFT, respectively.

To address \textbf{Q1}, we visualize the individual embeddings in a two-dimensional space by applying t-SNE \citep{van2008visualizing} for dimensionality reduction. Prior studies have investigated travel behavior by examining trip entropy \citep{goulet2017measuring, huang2019exploring}. To evaluate whether individual embeddings capture routing preferences, we assign a label to each individual based on their route-choice entropy (RCE). Specifically, we model route choice as a tuple consisting of the upstream link, downstream link, and departure time (e.g., a time period within a day). The RCE is computed as follows: 
\begin{equation}
H = -\sum_{i=1}^{n} p_i \log(p_i),
\label{eq:route_choice_entropy}
\end{equation}
where \( n \) is the number of different tuples and \( p_i \) is the probability that tuple \( i \) is chosen. The logarithm base used can vary depending on context (e.g., base 2 for binary entropy, or natural logarithm for information measured in nats). Intuitively, a higher RCE indicates a more unpredictable routing preference, whereas a lower RCE suggests that the individual follows a more consistent routing pattern.

\begin{figure}[bpht]
    \centering
\begin{subfigure}[b]{0.45\textwidth}
        \centering
        \includegraphics[width=\textwidth]{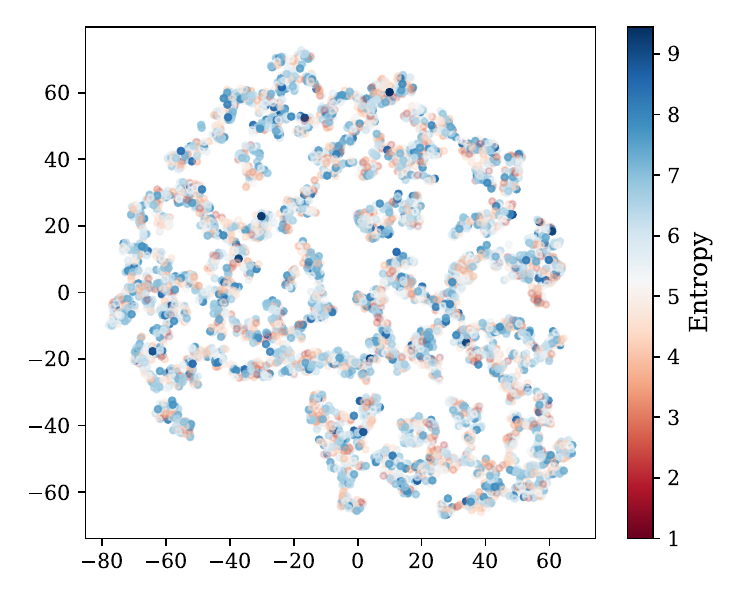}
        \caption{Initialization phase}
        \label{fig:individual_ini}
    \end{subfigure}
    \hfill
    \begin{subfigure}[b]{0.45\textwidth}
        \centering
        \includegraphics[width=\textwidth]{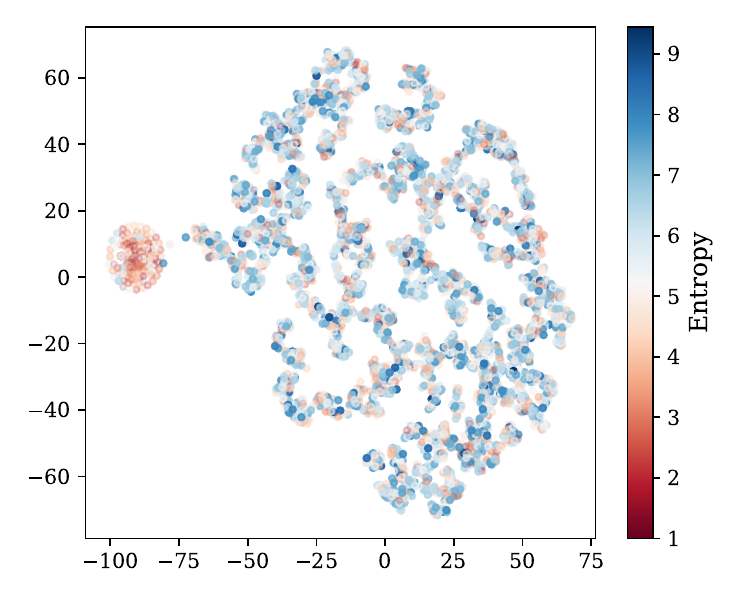}
        \caption{pre-trained phase}
        \label{fig:individual_pre}
    \end{subfigure}

    \begin{subfigure}[b]{0.45\textwidth}
        \centering
        \includegraphics[width=\textwidth]{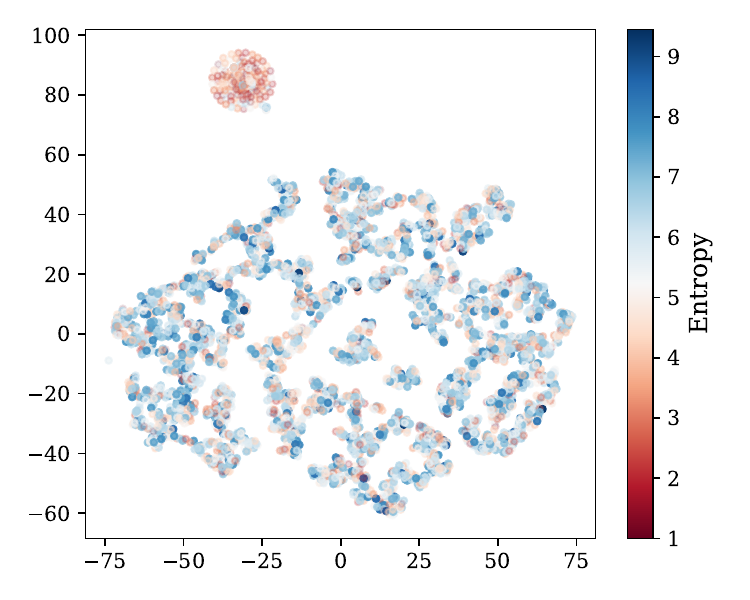}
        \caption{DPO phase}
        \label{fig:individual_dpo}
    \end{subfigure}
    \hfill
    \begin{subfigure}[b]{0.45\textwidth}
        \centering
        \includegraphics[width=\textwidth]{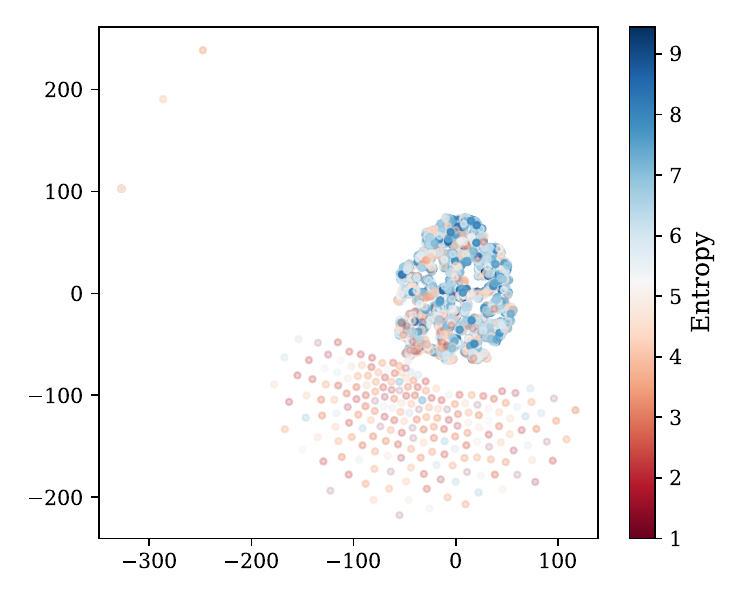}
        \caption{RMFT phase}
        \label{fig:individual_rlhf}
    \end{subfigure}
\caption{\textbf{Visualization of individual embeddings.} \small{Individual embeddings are projected into a two-dimensional space using t-SNE and colored according to route-choice entropy. Both the pretrained and fine-tuned models are able to identify distinct individual clusters. Notably, the RMFT-based fine-tuning results in a more dispersed distribution among individuals with low route-choice entropy, suggesting that RMFT enhances the model’s ability to capture subtle navigation differences, even for individuals with highly consistent travel routines.}}

    \label{fig:ub}
\end{figure}

As illustrated in Figure~\ref{fig:ub}, the individual embeddings in a two-dimensional space exhibit interesting patterns at various training stages. Initially, the embeddings from the initialized model appear disorganized, reflecting the absence of learned information. After Phase 1, the embeddings from the pre-trained model distinctly form two clusters corresponding to lower and higher entropy. This separation underscores the ability of the individual embeddings to capture essential information about travel regularity.

More interestingly, while the clustering pattern remains largely unchanged after fine-tuning with the DPO scheme, the proposed scheme shows a significant evolution in its clustering (see Figure~\ref{fig:individual_rlhf}). Compared to other stages, the low-entropy cluster from the proposed RMFT becomes more dispersed, suggesting an enhanced capability of the model to differentiate individuals with regular travel patterns. Conversely, the high-entropy cluster appears more condensed, indicating that the model effectively isolates irregular travel behavior, treating it akin to noise. These observations suggest that the RMFT significantly refines the model's ability to discern and categorize individual behaviors based on their travel regularity.

\begin{figure}[htbp]
    \centering
\begin{subfigure}[b]{0.46\textwidth}
        \centering
        \includegraphics[width=\textwidth]{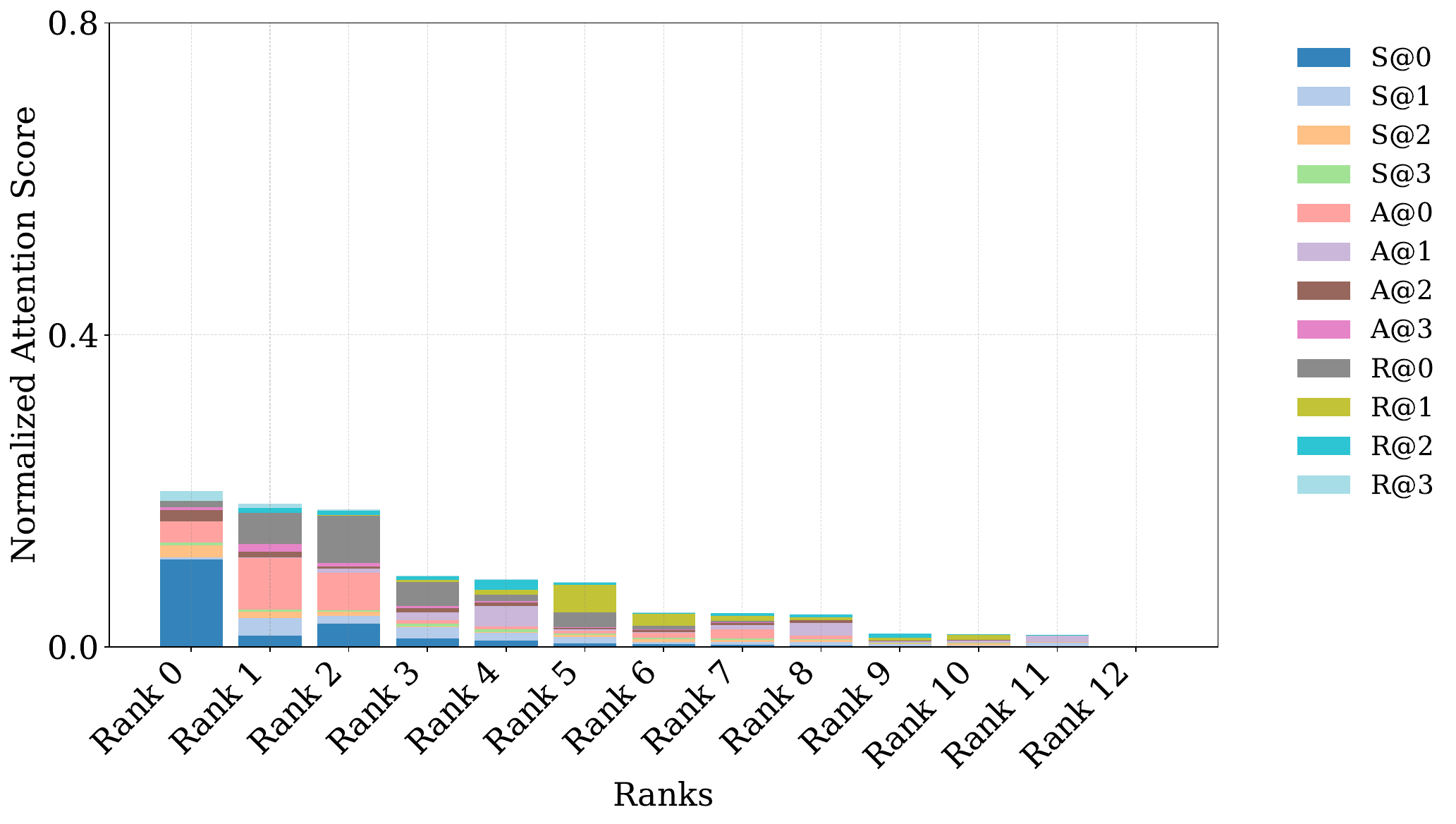}
        \caption{Initialization phase}
        \label{fig:figure1}
    \end{subfigure}
    \hfill
    \begin{subfigure}[b]{0.46\textwidth}
        \centering
        \includegraphics[width=\textwidth]{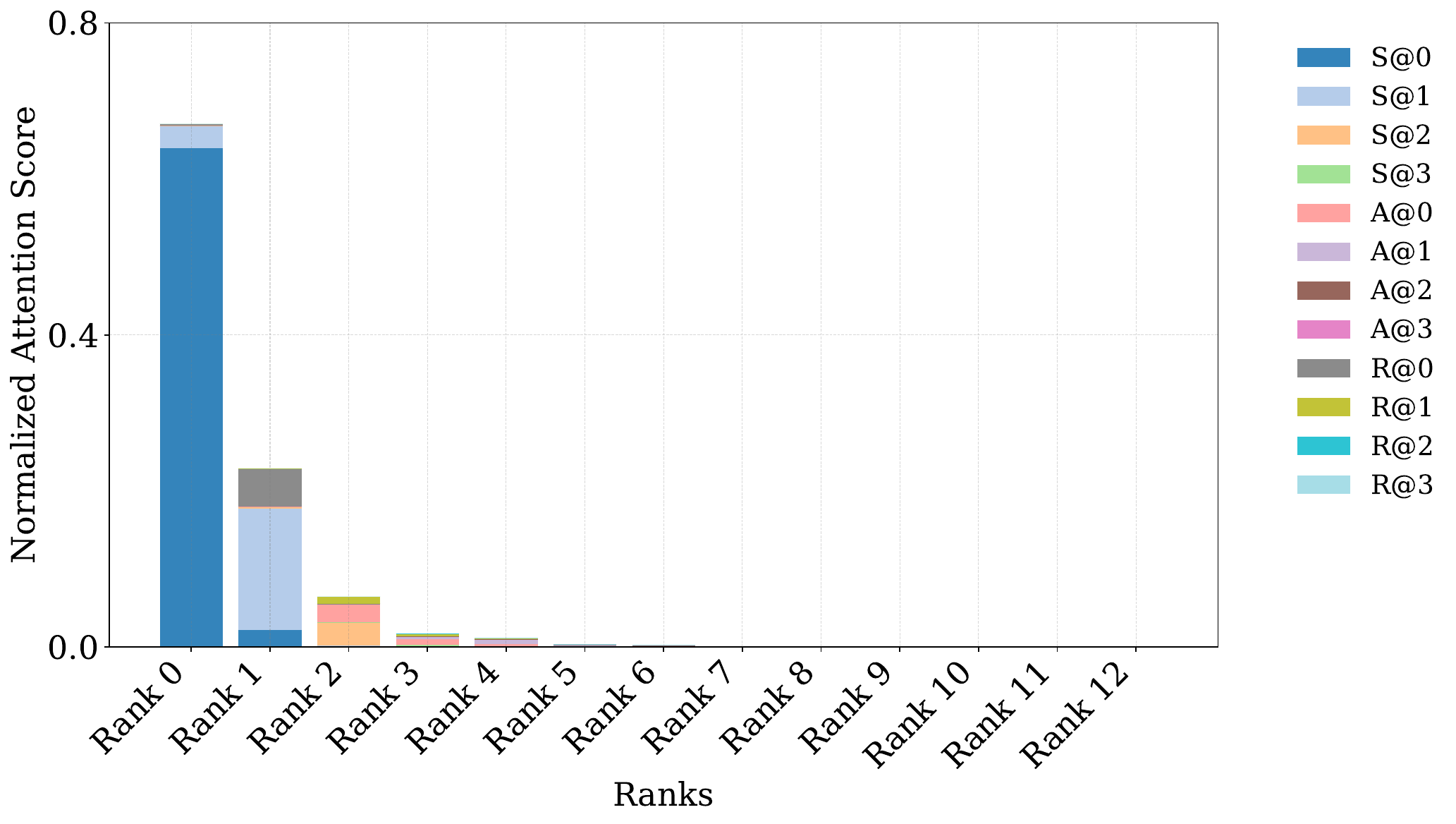}
        \caption{pre-trained phase}
        \label{fig:figure2}
    \end{subfigure}

    \begin{subfigure}[b]{0.46\textwidth}
        \centering
        \includegraphics[width=\textwidth]{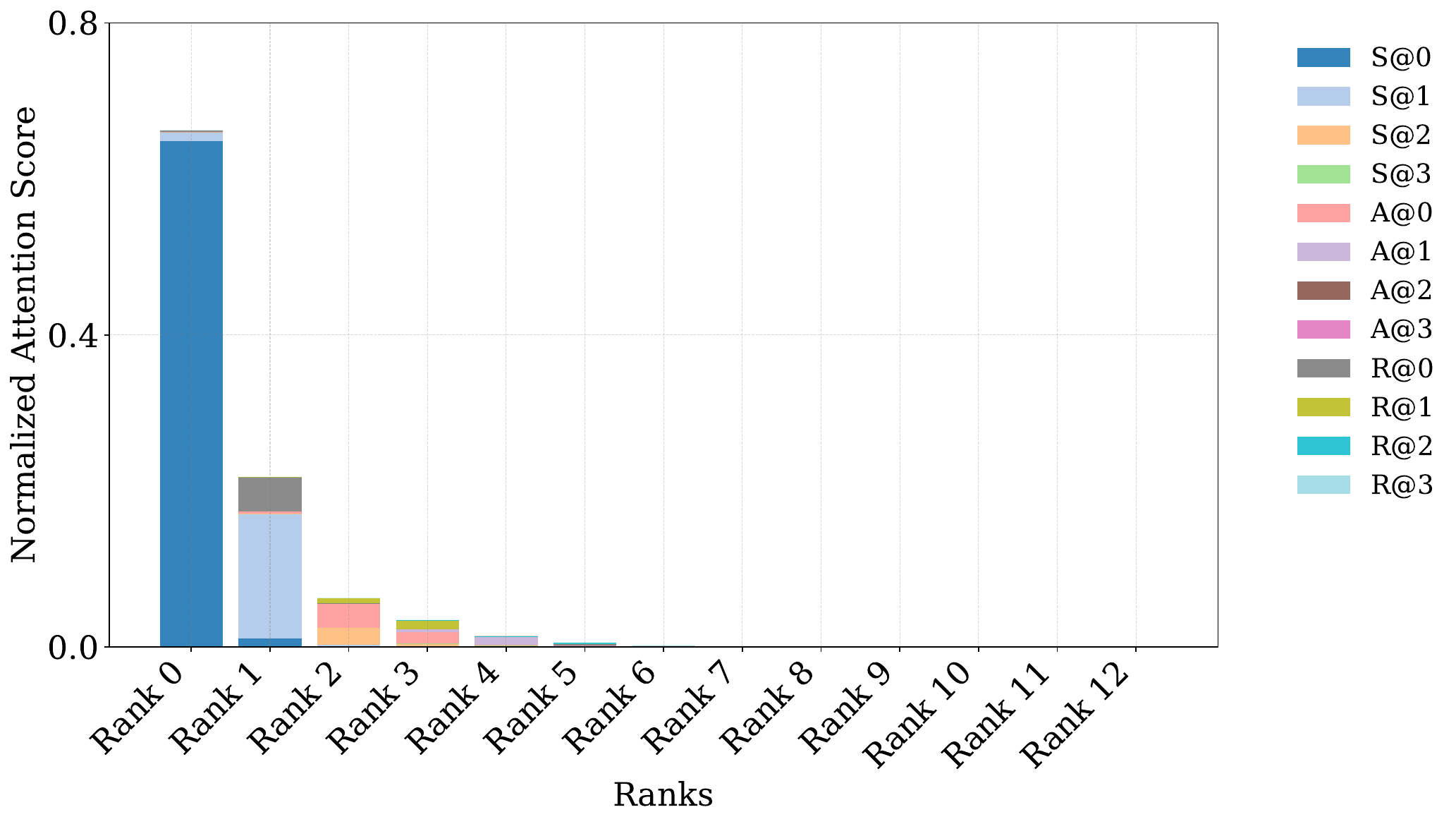}
        \caption{DPO phase}
        \label{fig:figure3}
    \end{subfigure}
    \hfill
    \begin{subfigure}[b]{0.46\textwidth}
        \centering
        \includegraphics[width=\textwidth]{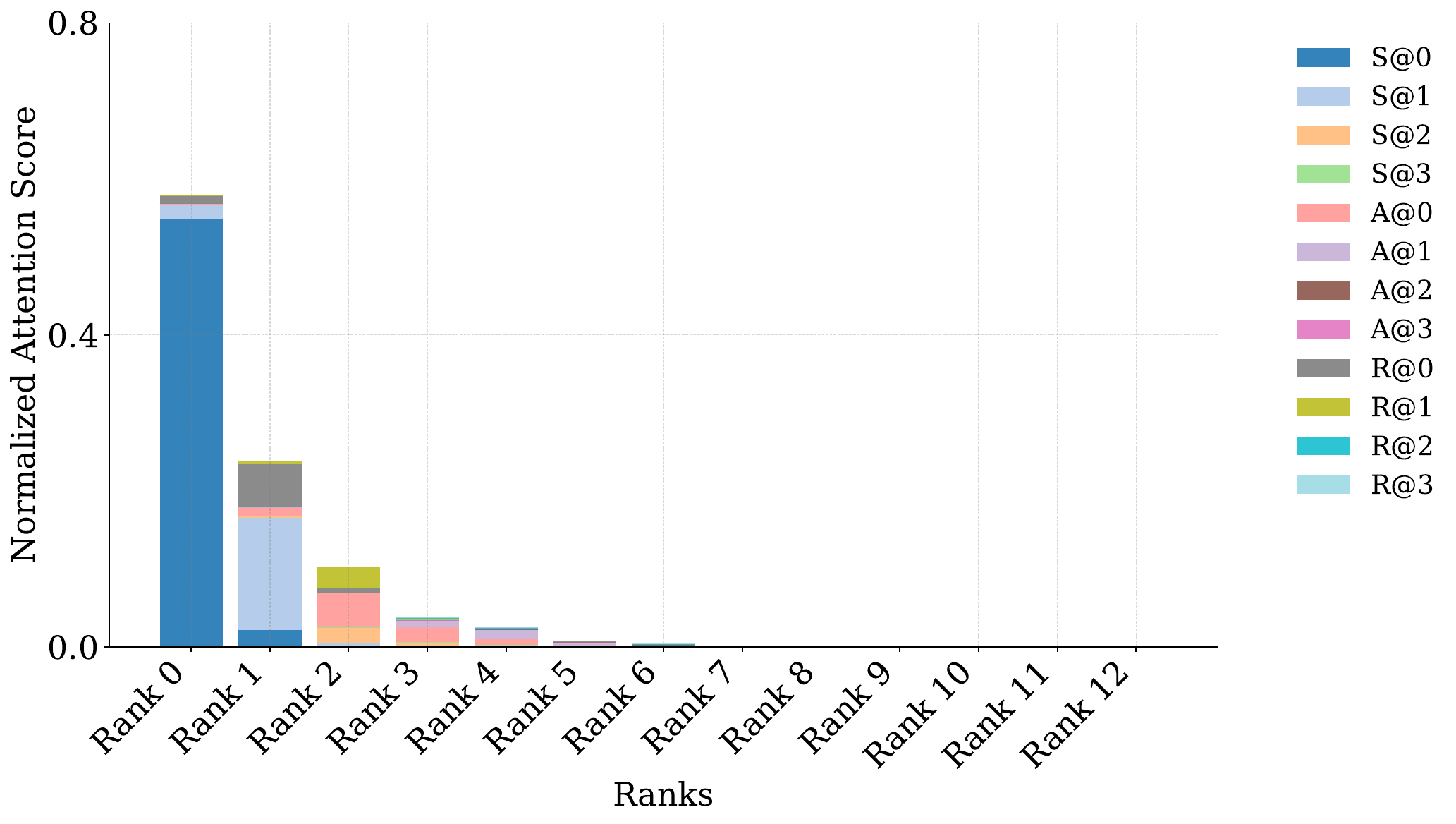}
        \caption{RMFT phase}
        \label{fig:figure4}
    \end{subfigure}

\caption{\textbf{Normalized attention scores for different tokens at various relative positions.}\small{ Here each token (e.g., state, action and return-to-go) is represented as a token type concatenated with the relative time step at which it occurs during the generation process.}}

    \label{fig:att_score}
\end{figure}

In the next step, we analyze how the model leverages input information to enhance the generation tasks (i.e., \textbf{Q2} and \textbf{Q3}). To explore these aspects, we first collect and rank attention scores for various tokens at different relative time steps in the generation process. As depicted in Figure~\ref{fig:att_score}, the notations \(S@t\), \(R@t\), and \(A@t\) refer to the state, return-to-go, and action tokens, respectively. Each token occurs at the \(t\)th relative time step in generating the future action, where a smaller \(t\) value indicates a closer location to the current generation output. Take the generation of the \(T\)th action as an example, \(S@0\) refers to the state token immediately before the generation of the \(T\)th action, and \(S@1\) represents the state token one step prior, used in generating the \(T\)th action.

Note that we limit the analysis to the top 12 ranked attention scores for clarity in the presentation. It is observed that, compared to the initialization, the models after different training phases exhibit a more distinct pattern in assigning attention scores. Specifically, the state token immediately preceding each generation step (i.e. \(S@0\)) consistently receives the highest attention score most of the time. This makes sense as the most immediate observation often provides the most relevant information for decision-making. Besides, this observation aligns with the formulation of trajectory generation as a partially observable Markov decision process, where future actions are primarily determined by the immediate state observation.
Furthermore, the proposed scheme leads to a more diverse distribution of attention scores compared to the other two schemes. This observation aligns with previous findings, indicating that the model fine-tuned with an explicit reward model tends to utilize more nuanced information effectively.
Finally, we delve deeper to examine specific token combination patterns during the generation process. At each generation step, three types of input tokens—return-to-go token, state token, and action token—are fed into the model as the latest information. By analyzing how the model reorders the importance of these tokens at each step, we aim to uncover deeper insights into the mechanism of the proposed model's generation scheme.

In our analysis, we focus on regular token combination patterns according to the relative time steps and token types. First, we reorder the preceding tokens for each trajectory generation based on their attention scores. Then, we concatenate every three adjacent tokens, treating them as the reorganized token combinations generated by the model.

\begin{figure}[ht]
\centering
\includegraphics[width=1\textwidth]{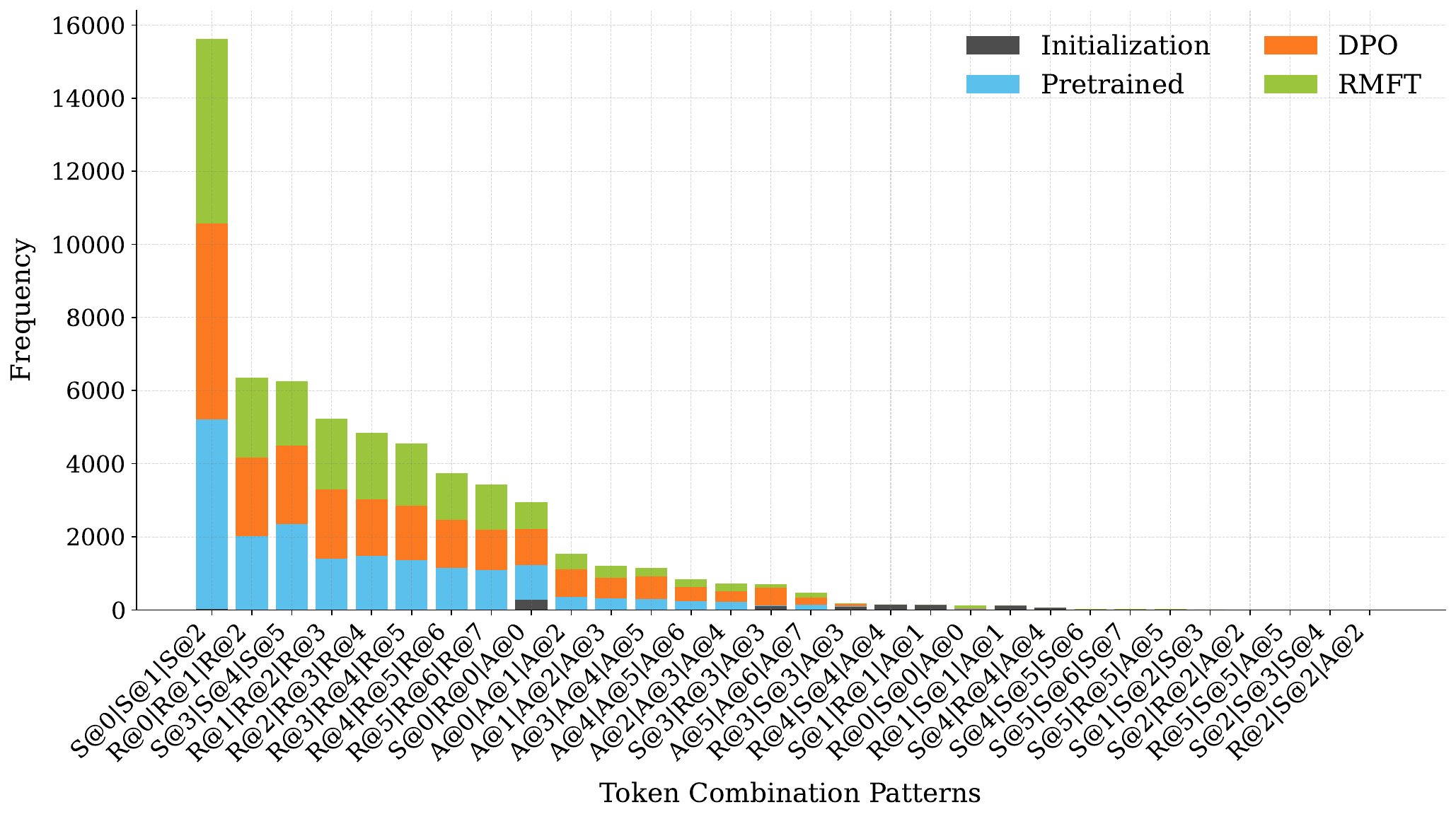}
\caption{\textbf{Frequency of selected regular token combinations.} \small{A higher frequency for a specific token combination pattern indicates that the pattern occurs more often throughout the entire generation process.}}
\label{fig:token_comb}
\end{figure}

As illustrated in Figure~\ref{fig:token_comb}, we quantify the frequency of occurrence of each regular token combination. The analysis reveals that post-training, the model shifts away from favoring the standard input token combination \(R@t|S@t|A@t\), instead learning to value a variety of combinations deliberatively. Notably, the combination \(S@0|S@1|S@2\) is most prevalent, indicating that the model has learned to prioritize the most recent state observations for subsequent generation steps. Furthermore, token combinations generally appear in chronological sequence, such as \(S@t|S@t+1|S@t+2\), \(R@t|R@t+1|R@t+2\), and \(A@t|A@t+1|A@t+2\).
Such phenomena illustrate that the importance of different tokens is reassessed in our auto-regressive decision-making scheme, with a particular emphasis on maintaining temporal sequential order. This behavior demonstrates the model's ability to discern and preserve temporal relationships critical for effective generation.

\section{Discussion}\label{sec3}
In this study, we introduce \textit{TrajGPT-R}, an innovative framework for urban mobility trajectory generation, engineered to overcome significant challenges inherent in existing methods. 
In the first phase, we focus on understanding trajectories by employing a transformer-based generative model and constructing a trajectory-wise reward model. Specifically, we frame the trajectory generation task as an offline RL problem, which benefits from a significant reduction of the vocabulary space during tokenization. To construct an informative reward model, IRL (Inverse Reinforcement Learning)--based reward modeling is introduced to learn trajectory-wise reward signals, enhancing our understanding of individual trajectory preferences. In the second phase, the pre-trained generative model (TrajGPT) is fine-tuned using the constructed reward model. This approach addresses two significant limitations in RL-based autoregressive generation schemes for urban mobility trajectory data: (1) the lack of consideration for long-term credit assignment, and (2) the presence of sparse reward signal information.

Through rigorous experimentation and comparisons with diverse baseline models across multiple large-scale trajectory datasets, our framework consistently demonstrated superior performance in both reliability and diversity metrics. Beyond performance metrics, we conducted a comprehensive interpretability analysis from three distinct perspectives, shedding light on the underlying mechanisms of our trajectory generation framework. This multifaceted analysis reinforces the robustness and practicality of our method, offering intuitive insights that bridge theoretical contributions and real-world applicability.

While our framework achieves state-of-the-art results, it also highlights several promising directions for future research that could further elevate the field:
\begin{enumerate}
    \item \textcolor{red}{In this study, we view the action tokenization as a practical trade-off between expressiveness and scalability. Extending the framework to support adaptive or hierarchical action vocabularies for high-degree intersections is a promising direction for future work.}
    \item \textcolor{red}{Although the experimental evaluation in this work focuses on vehicle trajectories, the proposed framework is not inherently limited to a single transportation mode. Extending TrajGPT-R to multi-modal mobility data—such as walking, cycling, and mixed-mode trajectories (e.g., GeoLife) by incorporating mode-aware state representations and reward functions is an important avenue for future research. Such extensions would further align the framework with real-world urban mobility systems and broaden its applicability.}
    \item Compounding errors remain a challenge in the autoregressive generation scheme, particularly for the generation of extremely long trajectories. Efforts to mitigate these errors could include advanced techniques such as improved sampling strategies, enhanced training procedures that better mimic the inference conditions, or novel architectural modifications that increase the model's robustness to early errors.
    \item While the framework explicitly models individual preferences through individual embeddings, these embeddings are not fully exploited during the testing phase, particularly concerning their transferability. Future research could explore methods to enhance the utilization of learned embeddings to achieve generalization across different scenarios. 
    \item Unlike traditional RLHF, which may directly incorporate human insights, our approach establishes a fully data-driven reward model. This method introduces potential biases due to the absence of explicit human intervention. A promising direction for future research involves iteratively updating the reward model based on individual preferences. Such updates would allow the model to adapt more effectively to personalized applications, enhancing functionalities like personalized navigation.
\end{enumerate}

\section*{Acknowledgements}
This research was supported by JSPS KAKENHI Grant Number JP25K21264 and JP24K02996.

\bibliographystyle{plainnat}
\bibliography{reference}
\newpage
\appendix
 
\section{Appendix}\label{secA1}
\subsection{Metric}
\noindent \textbf{1. Jaccard Similarity (Jac):}
\[
  \text{Jac}(A, B) = \frac{|A \cap B|}{|A \cup B|},
\]
where \( A \) and \( B \) are sets of elements (e.g., links of each trajectory).

\noindent \textbf{2. Cosine Similarity (Cos):}
\[
  \text{Cos}(\mathbf{a}, \mathbf{b}) = \frac{\mathbf{a} \cdot \mathbf{b}}{\|\mathbf{a}\| \|\mathbf{b}\|},
\]
where \( \mathbf{a} \) and \( \mathbf{b} \) are vector representations of trajectories.

\noindent \textbf{3. BLEU (Bilingual Evaluation Understudy):} \citep{papineni2002bleu}. BLEU (Bilingual Evaluation Understudy) is traditionally used to evaluate the quality of text generated by machine translation systems. We adapt it to measure trajectory diversity due to the similarity of the trajectory and text.
BLEU score calculations involve modified n-gram precisions and a brevity penalty, formulated as:
    \[
    \text{BLEU} = \text{BP} \cdot \exp\left(\sum_{k=1}^K w_k \log p_k\right),
    \]

where \( p_k \) is the precision for k-grams, computed as the ratio of the number of matching k-grams in \( t_i \) to the total number of k-grams in \( t_i \). \( w_k \) is the weight for each k-gram precision, typically uniform across different k values. The brevity penalty (BP) is calculated as:
\[
\text{BP} = \begin{cases} 
1 & \text{if } c > r \\
\exp(1 - r/c) & \text{if } c \leq r
\end{cases}
\]
where \( c \) is the length of the candidate sequence and \( r \) is the effective reference corpus length.

\noindent \textbf{4. Jessen-divergence of Link Distribution (L-JSD):}
\[
  L\text{-JSD}(P \parallel Q) = \frac{1}{2} \left[ D(P \parallel M) + D(Q \parallel M) \right],
\]
where \(P\) and \(Q\) are the probabilities of link segments in the generated data and the ground-truth data. \( M = \frac{1}{2}(P + Q) \) and \( D \) is the Kullback-Leibler divergence. This metric is used to measure the link distribution proximity between the generated and the real data.

\noindent \textbf{5. Jessen-divergence of Connection Distribution (C-JSD):}
\[
  C\text{-JSD}(P \parallel Q) = \frac{1}{2} \left[ D(P \parallel M) + D(Q \parallel M) \right].
\]
This metric is similar to L-JSD but applied to connection distributions.

\noindent \textbf{6. Unigram Entropy (UE):}
\[
  UE = - \sum_{i \in L} p_i \log_2(p_i),
\]
where \(L\) represents the set of all unique links in the dataset, and \( p_i \) is the probability of the \(i\)-th link ID occurring in the generated trajectories. This metric quantifies the diversity of individual links, reflecting their variety at the most granular level.

\noindent \textbf{7. Bigram Entropy (BE):}
\[
  BE = - \sum_{(i,j) \in C} p_{ij} \log_2(p_{ij}),
\]
where \(C\) denotes the set of all unique consecutive link pairs (connections) in the dataset, and \( p_{ij} \) is the probability of the pair \( (i, j) \) occurring in the generated trajectories. This metric evaluates the diversity of transitions between consecutive links, providing insight into the local structural variety.

\subsection{Model Configurations}\label{secA2} 

This appendix provides the configuration details of the deep learning models implemented using PyTorch for the experiments discussed in the main document. The models are designed to cater to different environments, namely Toyota, T-Drive and Porto datasets. GPT2Model, a variant of the GPT-2 architecture, is served as backbone with a specific configuration for different data.

\subsection*{Environment-specific Configurations}
\noindent \textbf{1. Toyota dataset configuration}
\begin{itemize}
    \item \textbf{Embeddings:} Various embeddings are employed to encode different types of inputs:
        \begin{itemize}
            \item \textbf{Link, Origin, Destination:} Embedded using separate embeddings with a vocabulary size of 262144.
            \item \textbf{Action, Departure Time:} Action dimensions and departure times are embedded with their respective sizes.
            \item \textbf{Speed:} Embedded using an embedding layer designed for a range of 120 different speeds.
        \end{itemize}
    \item \textbf{Layer Normalization:} Applied post-embedding to stabilize the learning process.
    \item \textbf{Predictive Outputs:} Includes prediction of actions and optionally states and returns, facilitated by linear transformations and activation functions.
\end{itemize}

\noindent \textbf{2. T-Drive Dataset Configuration}
Similar to the Toyota configuration with adjustments to embedding sizes for Link, Origin, and Destination, each reduced to a vocabulary size of 16384.


\noindent \textbf{3. Porto Dataset Configuration}
Adapted embedding sizes for Link, Origin, and Destination, reflecting the smaller geographic scope and dataset size with a vocabulary size of 5524.

\subsection*{Common Features across Environments}
\begin{itemize}
    \item \textbf{Individual Embeddings:} We use word embedding with the actual number of individuals as the vocabulary size.
    \item \textbf{Timestep Embedding:} We use word embedding with a maximum trajectory length as the vocabulary size.
\end{itemize}

\subsection*{Model Training Configurations}

\begin{table}[h!]
\centering
\caption{Summary of Training Configurations}
\label{tab:training_configurations}
\begin{tabular}{p{3.5cm} >{\centering\arraybackslash}p{2cm} >{\centering\arraybackslash}p{2cm} >{\centering\arraybackslash}p{2cm}}
\toprule
\textbf{Parameter}        & \textbf{Toyota} & \textbf{T-Drive} & \textbf{Porto} \\
\midrule
Weight Decay             & 0.05            & 0.02             & 0.05          \\
Embedding Dimension      & 512             & 256              & 256           \\
Learning Rate            & 0.0005          & 0.0005           & 0.0005        \\
Sub-sample Length        & 64              & 12               & 64            \\
Batch Size               & 64              & 64               & 128           \\
Attention Layers         & 2               & 2                & 3             \\
\bottomrule
\end{tabular}
\end{table}

{
\subsection*{Implementation Details of the IRL-Based Reward Model}
\label{app:irl_reward}
This appendix provides the technical details of the inverse reinforcement learning (IRL)–based reward model used in Phase~1. We describe the model architecture, the definitions of the Base Value Estimator (BVE) and Preference Value Estimator (PVE), and the associated training assumptions.

\textbf{1. State and Action Representation}

At each time step $t$, the state $s_t$ is represented by a discrete feature vector consisting of:
\begin{itemize}
    \item current link ID,
    \item origin link ID,
    \item destination link ID,
    \item departure-time bin,
    \item user ID.
\end{itemize}
All attributes are treated as categorical variables and mapped to continuous embeddings. The action space corresponds to selecting one of the downstream links from the current link, yielding a discrete action dimension $|\mathcal{A}|$.

\textbf{2. Reward Model Architecture}

The reward model is implemented as a critic network parameterized by $\phi$, which outputs an action-value estimate decomposed into two components: a Base Value Estimator (BVE) and a Preference Value Estimator (PVE).
We employ independent embedding layers for link ID, origin link, destination link, departure-time bin, and user ID. All embeddings share the same hidden dimension $d$.

\paragraph{Base Value Estimator (BVE)}
The BVE captures the general spatiotemporal desirability of state--action pairs independent of individual preferences. Specifically, the embeddings of the origin link, destination link, current link, and departure-time bin are concatenated and projected into a state embedding:
\[
h_t = \sigma\left( W_s \,[e_o, e_d, e_{\text{link}}, e_{\text{depart}}] \right),
\]
where $\sigma(\cdot)$ denotes a LeakyReLU activation.

The BVE output is then computed as:
\[
\text{BVE}(s_t,a_t) = W_{\text{bve}} h_t \in \mathbb{R}^{|\mathcal{A}|},
\]
producing an action-value vector over downstream-link choices.

\paragraph{Preference Value Estimator (PVE)}
The PVE models trajectory-wise preference biases associated with individual users. A preference embedding is constructed by summing the embeddings of origin, destination, departure time, and user ID:
\[
p_t = e_o + e_d + e_{\text{depart}} + e_{\text{user}}.
\]

The PVE output is defined as:
\[
\text{PVE}(s_t,a_t) = W_{\text{pve}} p_t \in \mathbb{R}^{|\mathcal{A}|},
\]
which introduces a preference-dependent bias over actions.

\paragraph{Final Reward Signal}
The reward model outputs an action-value estimate by combining the two components:
\[
Q_\phi(s_t,a_t) = \text{BVE}(s_t,a_t) + \text{PVE}(s_t,a_t).
\]
Trajectory-wise rewards are computed by aggregating action-values along the trajectory and are subsequently used to derive advantage estimates via generalized advantage estimation (GAE).

\textbf{3. IRL Training Details}

The reward model is trained using an inverse reinforcement learning objective with preference-labeled trajectory data. The discount factor $\gamma = 0.9$ and GAE parameter $\gamma\lambda = 0.95$ follow standard practice and are fixed across all experiments. Transition dynamics are implicitly captured by the empirical trajectory data, and no explicit parametric transition model is estimated.

}

{
\subsection*{Inference Time Cost Analysis}
\label{app:inference_time}
To evaluate the computational efficiency of different trajectory generation paradigms, we compare the average inference time required to generate a single trajectory across representative model families and datasets. All models are evaluated under the same hardware and software configuration, and the reported time corresponds to the average inference latency per trajectory. 
 
Table~\ref{tab:inference_time} summarizes the results. As expected, diffusion-based models incur a higher inference cost due to their iterative denoising process, while GAN- and VAE-based models benefit from single-pass generation. Autoregressive models exhibit moderate inference time with linear dependence on trajectory length, representing a trade-off between efficiency and expressive sequential modeling capability.

\begin{table}[htbp]
\centering
\caption{Inference time comparison for different trajectory generation models. Reported values indicate the average time (in milliseconds) required to generate a single trajectory.}
\label{tab:inference_time}
\begin{tabular}{l c}
\hline
\textbf{Model Type} & \textbf{Inference Time (ms)} \\
\hline
Autoregressive-based & 45.0 \\
Diffusion-based     & 80.0 \\
GAN-based           & 12.0 \\
VAE-based           & 15.0 \\
\hline
\end{tabular}
\end{table}
Overall, this comparison highlights the trade-off between inference efficiency and modeling flexibility. While autoregressive models are slower than fully parallel generative models, their inference cost remains manageable for offline large-scale urban simulation, which is the primary application setting of this work.
}

\subsection*{Ethics Statement}
This work complies with the ICLR Code of Ethics. The study leverages large-scale, anonymized mobility trajectory datasets for training and evaluation. All data were handled in accordance with strict privacy and data-use regulations. No personally identifiable information (PII) or user-level demographics (e.g., age, gender, home–work identifiers) were used, and no attempt was made to reconstruct or deanonymize individual behavior.  

The proposed models are designed for methodological advancement in trajectory generation and urban mobility analysis. While the research may inform applications such as transportation planning or disaster response, it does not involve human subjects or interventions, nor does it seek to predict or profile individuals. Potential risks of misuse, including surveillance or discriminatory practices, are acknowledged; to mitigate these, we restrict our work to anonymized data and emphasize aggregate-level evaluation.  

We believe this research contributes positively to society by providing tools that can support urban planning, traffic management, and resilience studies. All ethical, legal, and research integrity standards have been respected throughout this work.

\end{document}